\documentclass[journal]{IEEEtran}
\usepackage{graphicx}
\usepackage[english]{babel}
\usepackage{microtype}
\usepackage{caption}
\usepackage{placeins}  
\usepackage{tabularx}
\usepackage{booktabs}
\usepackage{multirow}
\usepackage{caption}
\usepackage{natbib}
\usepackage{url}
\usepackage{amsmath}
\usepackage{amssymb}
\ifCLASSINFOpdf
\else
\fi
\begin{document}

%
% paper title
% Titles are generally capitalized except for words such as a, an, and, as,
% at, but, by, for, in, nor, of, on, or, the, to and up, which are usually
% not capitalized unless they are the first or last word of the title.
% Linebreaks \\ can be used within to get better formatting as desired.
% Do not put math or special symbols in the title.
\title{MultiST: A Cross-Attention-Based Multimodal Model for Spatial Transcriptomics}
%
%
% author names and IEEE memberships
% note positions of commas and nonbreaking spaces ( ~ ) LaTeX will not break
% a structure at a ~ so this keeps an author's name from being broken across
% two lines.
% use \thanks{} to gain access to the first footnote area
% a separate \thanks must be used for each paragraph as LaTeX2e's \thanks
% was not built to handle multiple paragraphs
% %\author[1]{Wei Wang}
% \author[1]{Quoc-Toan Ly}
% \author[1,$\ast$]{Chong Yu}
% \author[1,$\ast$]{Jun Bai}

% \author{Wei~Wang, %~\IEEEmembership{Member,~IEEE,}
%         Quoc-Toan~Ly, %~\IEEEmembership{Fellow,~OSA,}
%         Chong~Yu\textsuperscript{*}, %~\IEEEmembership{Fellow,~OSA,}
%         and~Jun~Bai\textsuperscript{*}% <-this % stops a space
% \thanks{*Corresponding author. Email: xxx@xxx.edu
% % M. Shell was with the Department
% % of Electrical and Computer Engineering, Georgia Institute of Technology, Atlanta,
% % GA, 30332 USA e-mail: (see http://www.michaelshell.org/contact.html).
% }% <-this % stops a space
% \thanks{J. Doe and J. Doe are with Anonymous University.}% <-this % stops a space
% % \thanks{Manuscript received April 19, 2005; revised August 26, 2015.}
% }
\author{
Wei~Wang,
Quoc-Toan~Ly,
Chong~Yu\textsuperscript{*},
and~Jun~Bai\textsuperscript{*}%
\thanks{All authors are with the Department of Computer Science, University of Cincinnati,
Cincinnati, OH 45221, USA.}%
\thanks{*Corresponding authors. Emails: yuc5@ucmail.uc.edu; baiju@ucmail.uc.edu.}%
}

\maketitle

% As a general rule, do not put math, special symbols or citations
% in the abstract or keywords.
\begin{abstract}
% 100-200 words
% \textbf{Motivation:}
Spatial transcriptomics (ST) enables transcriptome-wide profiling while preserving the spatial context of tissues, offering unprecedented opportunities to study tissue organization and cell–cell interactions in situ. 
% However, inherent data sparsity, measurement noise, and the difficulty of integrating molecular profiles with histological information hinder accurate and interpretable downstream analyses. 
% Despite recent advances, existing methods often lack effective integration of histological morphology with molecular profiles, relying on shallow fusion strategies or omitting tissue images altogether, which limits their ability to resolve ambiguous spatial domain boundaries. Consequently, a computational framework that jointly models spatial, molecular, and morphological information to produce spatially coherent and biologically meaningful representations remains lacking.
% A computational framework that effectively combines spatial, molecular, and morphological information to yield spatially coherent and biologically meaningful representations remains lacking.
% \textbf{Results:}
% In this research, we propose MultiST, a multimodal framework that jointly models spatial topology, molecular expression, and tissue morphology to address these challenges. 
% MultiST employs a graph-based gene encoder with adversarial alignment to learn robust, spatially informed representations, together with color-normalized image encoders and a cross-attention fusion mechanism to capture molecular–morphological dependencies and refine spatial domain boundaries.
Despite recent advances, existing methods often lack effective integration of histological morphology with molecular profiles, relying on shallow fusion strategies or omitting tissue images altogether, which limits their ability to resolve ambiguous spatial domain boundaries. To address this challenge, we propose MultiST,
a unified multimodal framework that jointly models spatial topology, gene expression, and tissue morphology through cross-attention–based fusion. MultiST employs graph-based gene encoders with adversarial
alignment to learn robust spatial representations, while integrating color-normalized histological features to capture molecular–morphological dependencies and refine domain boundaries.
% In this research, we propose MultiST, a multimodal framework for precise and interpretable downstream analysis of spatial transcriptomics data. 
% The gene encoder employs graph-based embeddings with adversarial alignment for robust representation learning under sparse and noisy conditions, 
% while color-normalized image encoders and cross-attention fusion capture complementary molecular–morphological dependencies. 
% We evaluate the proposed method on 13 diverse ST datasets, MultiST consistently outperforms existing methods in clustering, pseudotime trajectory inference, and cell–cell interaction analysis. It produces spatially coherent and biologically interpretable multimodal representations of tissue architecture.
We evaluated the proposed method on 13 diverse ST datasets spanning two organs, including human brain cortex and breast cancer tissue. MultiST yields spatial domains with clearer and more coherent boundaries than existing methods, leading to more stable pseudotime trajectories and more
biologically interpretable cell–cell interaction patterns.
% We evaluated MultiST on 13 diverse ST datasets. 
% Across spatial domain identification, pseudotime trajectory inference, and cell–cell interaction analysis, MultiST consistently yields spatial domains with clearer and more coherent boundaries, leading to more reliable and biologically interpretable downstream results.
% \textbf{Availability and implementation:}
The MultiST framework and source code are available at  \url{https://github.com/LabJunBMI/MultiST.git}.
\end{abstract}

% Note that keywords are not normally used for peerreview papers.

% Applied computing →Bioinformatics; Computational biology; • Computing methodologies →Neural networks.

% For peer review papers, you can put extra information on the cover
% page as needed:
% \ifCLASSOPTIONpeerreview
% \begin{center} \bfseries EDICS Category: 3-BBND \end{center}
% \fi
%
% For peerreview papers, this IEEEtran command inserts a page break and
% creates the second title. It will be ignored for other modes.
\IEEEpeerreviewmaketitle

% \vspace{-5pt}
\section{Introduction}
Spatial transcriptomics (ST) is an emerging genomic technology that enables transcriptome-wide gene expression profiling while preserving the spatial organization of intact tissue sections~\citep{ST1,ST2}. Compared to single-cell RNA sequencing (scRNA-seq), ST does not require tissue dissociation, thereby avoiding stress responses, cell loss, and the disruption of spatial context~\citep{STvsscRNA,dissociation_artifacts}. This makes ST particularly suitable for studying spatial cell distribution, tissue heterogeneity, and intercellular communication within complex biological systems~\citep{ST_application,spatial_communication}.

% Despite its transformative potential in spatial biology, downstream analysis of ST data remains fraught with challenges. Under increasingly benchmark-driven scrutiny, persistent weaknesses are evident across core tasks: cell-type deconvolution, spatial domain detection, multi-slice/3D integration, and statistically sound inference under spatial dependence. Moreover, protocol choices and platform chemistry can dominate outcomes, introducing systematic biases that hinder method comparisons and biological interpretation ~\citet{li2023comprehensive}. In addition to these issues, high dropout rates, sparse high-dimensional expression matrices, heterogeneous spatial resolutions across platforms, and cell mixtures at the spot level further complicate modeling and validation~\citep{ST_noise, ST_platform_comparison}. Collectively, these challenges underscore the need for reproducible, metadata-aware pipelines and multimodal computational frameworks that jointly leverage gene expression, spatial coordinates, and histological context to deliver robust clinically meaningful insights~\citep{multimodal_ST}.

Despite its transformative potential, downstream analysis of ST data remains challenging, particularly in three core tasks: spatial domain identification, pseudotime trajectory inference, and cell–cell interaction (CCI).
In spatial domain identification, the key difficulty lies in balancing transcriptional homogeneity and spatial coherence, since spatially adjacent spots should exhibit similar expression profiles while preserving biologically meaningful boundaries.
However, transcriptomic transitions across anatomical interfaces, such as tumor–stroma boundaries, are often gradual rather than discrete, which complicates the distinction between neighboring domains.
Spots located in these transitional regions frequently co-express marker genes from both adjacent domains, and this overlap makes boundary assignment highly sensitive to algorithmic parameters instead of reflecting true biological continuity.
At the same time, histological features such as cell density and tissue texture encode complementary spatial information that could help resolve these ambiguous regions, but such information remains underexploited in most existing methods~\citep{clustering}.
Second, pseudotime trajectory inference relies on accurately defined spatial domains to reconstruct developmental trajectories. When domain boundaries are biologically inconsistent, inferred trajectories may become distorted, leading to incorrect ordering of cellular states.
Third, CCI analysis depends on reliable spatial neighborhoods to quantify ligand--receptor signaling patterns. Errors propagated from domain misclassification can obscure true communication pathways and amplify artifacts introduced by measurement noise, high dropout rates, mixed-cell spots, and tissue heterogeneity.

Because trajectory inference and CCI  typically rely on the identified domains to define developmental progressions and spatial neighborhoods for signaling estimation, any inaccuracies in boundary delineation can propagate downstream and reduce biological interpretability.
These challenges are further exacerbated by high dropout rates, mixed-cell spots, and tissue heterogeneity. Together, these issues highlight the need for a unified computational framework that capable of simultaneously modeling spatial topology, developmental dynamics, and distance-dependent cellular communication~\citep{challenge}.

%On one hand, gene expression matrices are inherently high-dimensional and sparse, often accompanied by substantial technical noise~\cite{ST_noise}. On the other hand, ST platforms differ in spatial resolution~\cite{ST_platform_comparison}, and individual spots frequently capture mixtures of multiple cell types~\cite{deconvolution_challenge}, further complicating downstream interpretation. These challenges highlight the need for computational frameworks capable of jointly modeling gene expression, spatial context, and histological features in a multimodal setting~\cite{multimodal_ST}.

To address these challenges, a range of computational methods have been developed. In the domain of statistical modeling, BayesSpace~\citep{BayesSpace} employs Bayesian inference and Markov Chain Monte Carlo (MCMC) sampling to achieve subspot-level spatial resolution by explicitly modeling neighborhood dependencies. This probabilistic formulation helps to smooth domain boundaries and reduce fragmentation compared with nonspatial clustering methods.
Deep learning–based approaches further extend this direction by leveraging graph structures and self-supervised objectives to learn spatially informed embeddings.
For example, SEDR~\citep{SEDR} integrates variational autoencoders (VAE) with graph convolutional networks (GCNs) to jointly encode spatial and transcriptional features, enabling more coherent domain delineation. STAGATE~\citep{STAGATE} applies graph attention networks to capture localized spatial dependencies and adaptively weight neighboring spots. SpaGCN~\citep{SpaGCN} incorporates both spatial coordinates and histological image features through graph construction, thereby enhancing the interpretability of the detected domains.
To further address the continuity of developmental trajectories, SpaceFlow~\citep{SpaceFlow} introduces spatial regularization to enforce smooth transitions along inferred gradients, improving pseudotemporal reconstruction under sparse and noisy expression conditions.
Meanwhile, conST~\citep{conST} adopts multimodal contrastive learning to align gene expression, spatial coordinates, and histological images in a unified latent space, providing a foundation for more spatially constrained analyses such as domain segmentation and communication inference.
Finally, Seurat~\citep{Seurat}, although originally developed for scRNA-seq, remains a widely used baseline for ST clustering and data integration via its shared nearest neighbor (SNN) graph construction.
Overall, these studies represent progressive efforts to achieve spatially coherent, noise-tolerant, and biologically interpretable representations of spatial transcriptomic data.

Although these methods have advanced various aspects of ST analysis, several important limitations remain.
First, most existing approaches lack effective multimodal integration. Some models include histological images, but they usually use shallow fusion strategies such as simple feature concatenation or separate encoders. These approaches cannot fully capture the relationships between tissue morphology and gene expression.
Many other methods rely only on gene expression and spatial coordinates, completely ignoring the structural information contained in tissue images.
In fact, including tissue images is important because morphological features help reveal cell neighborhoods, tissue organization, and local microenvironments that cannot be inferred from expression data alone. Without such information, the learned spatial representations may not match the true anatomical structures, leading to blurred boundaries or incorrect domain assignments~\citep{HE_image}. 
Second, although some models incorporate generative components, these modules are primarily designed for embedding reconstruction rather than for improving the underlying data quality.
% Another limitation is that models  include generative components, but these modules are mainly used for embedding reconstruction rather than improving data quality.
As a result, under sparse or noisy data conditions, these methods often show limited robustness and produce unstable clustering results, which can propagate errors into downstream analyses.

To bridge these gaps, we proposed MultiST, a novel multimodal spatial transcriptomics framework that jointly models gene expression, spatial coordinates, and histological image features. The goal is to achieve both computational accuracy and biological interpretability.

% Our framework is designed not only to achieve more accurate computational results, but also to enhance the biological interpretability of spatial patterns and reveal cell–cell interactions.

Our main contributions are as follows:
\vspace{-0.2em}
\begin{itemize}

\item \textbf{Cross-modal representation learning beyond shallow fusion.}
A cross-attention mechanism aligns histological morphology with gene expression, yielding spatially coherent and biologically meaningful representations.

\item \textbf{Robust graph-based encoding with self-supervised masking.}
An additive expression masking strategy enhances robustness to dropout, sparsity, and mixed-cell spots, leading to more stable spatial embeddings.

\item \textbf{Generative latent refinement via hybrid GAN--Fisher MMD.}
A hybrid GAN--Fisher MMD module regularizes the latent space and improves data quality, mitigating instability in clustering and trajectory inference.

\item \textbf{Stain-invariant image encoding for consistent multimodal fusion.}
Color-normalized CLIP-ViT features with spatial smoothing reduce staining variability and provide reliable morphological cues.

\item \textbf{Unified latent space for downstream biological analysis.}
The refined embedding supports accurate spatial domain segmentation, pseudotime reconstruction, and ligand--receptor CCI analysis.

\end{itemize}

\section{Methods}\label{sec2}
% \begin{figure*}[htbp]
%     \centering
%     \includegraphics[width=\textwidth]{image/new_frame.pdf}
%     \caption{\textbf{Overview of the MultiST framework.}    
% The proposed MultiST model integrates gene expression and histological image modalities to generate robust spatially informed embeddings for downstream analysis.
% \textbf{Gene modality (top)}: Input spot-level gene expression profiles are processed with masking and a graph-based encoder that incorporates spatial adjacency from K-nearest neighbors (KNN). A graph convolutional network (GCN) extracts spatial-aware features, followed by dual-path decoding for expression and graph reconstruction. A generative adversarial network (GAN) with Maximum Mean Discrepancy (MMD) regularization enforces latent distribution alignment, while deep embedding clustering (DEC) refines cluster compactness.
% \textbf{Image modality (bottom)}: H\&E-stained tissue images are preprocessed via color normalization, spot-aligned patch extraction, and feature encoding with a pretrained vision transformer. KNN-based smoothing further enhances spatial coherence of image embeddings.
% \textbf{Cross-modal fusion (right)}: Gene and image embeddings are integrated through a cross-attention mechanism, enabling fine-grained alignment of expression and morphological cues. The fused embeddings undergo label diffusion for refinement and are applied to multiple downstream tasks, including spatial clustering, pseudotime trajectory inference, and CCI analysis.}
%     \label{fig:framework}
% \end{figure*}
\begin{figure*}[htbp]
    \centering
    \includegraphics[width=\textwidth]{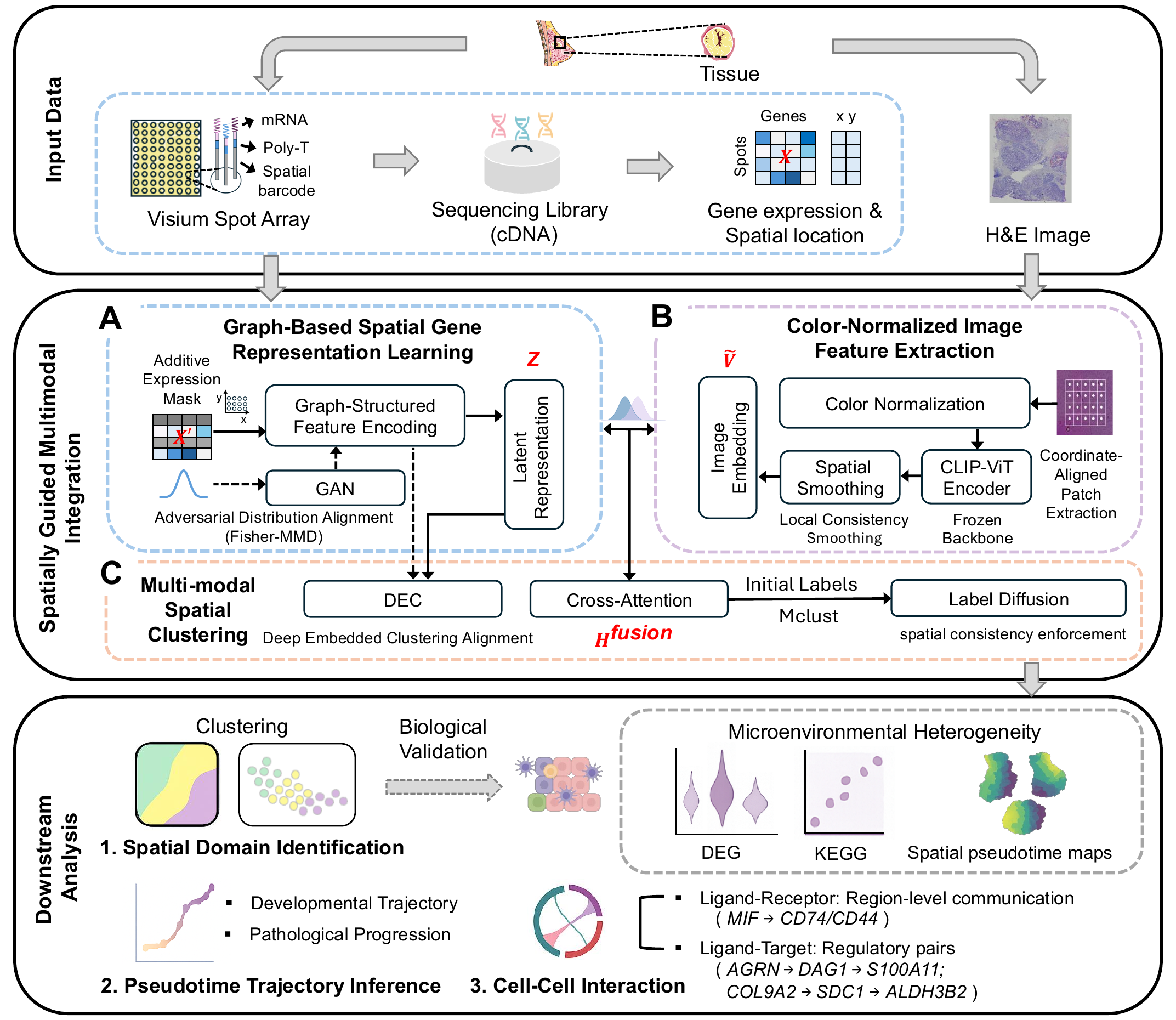}
    \vspace{-18pt}
    \caption{\textbf{The MultiST framework for spatially guided multimodal integration and biological discovery.}   
    Top: MultiST takes spatial transcriptomics data as input, including H\&E-stained tissue images and matched gene expression matrices with spatial barcodes. Middle: The Spatially Guided Multimodal Integration module models molecular and histological modalities via two parallel encoders. Latent representations are fused through cross-attention, with label diffusion enforcing spatial coherence. Bottom: The refined multimodal representations support downstream analyses, including spatial domain identification, pseudotime trajectory inference, and cell--cell interaction (CCI).
}
    \label{fig:framework}
    \vspace{-15pt}
\end{figure*}
    
% The aim of our propose method is to accelerate the novel bio-dispersion with novel AI based model. We proposed a novel MultiST model that integrates gene expression and histological image modalities to generate robust spatially informed embeddings for downstream analysis. Through effective downstream analysis, we proposed the method for the elimination of new biomarkers. 

% The proposed MultiST (Fig.\ref{fig:framework}), a unified dual-modal framework for spatial transcriptomics that jointly learns from gene expression and histological images. MultiST generates biologically informative spot-level embeddings to support downstream tasks such as spatial clustering, trajectory inference, and CCI analysis. It comprises three key components: a graph-based gene encoder, a color-normalized image encoder using a pretrained vision transformer, and a cross-attention fusion module that integrates multimodal features. 

% The aim of our proposed method is to advance spatial transcriptomics analysis through an AI-based multimodal framework. 
% We proposed MultiST, a unified dual-modal model that integrates gene expression and histological image information to learn robust, spatially informed embeddings for downstream analysis. 
% Through effective downstream analyses, MultiST contributes to the discovery of new biomarkers.

% The aim of this study is 
We advance spatial transcriptomics analysis by developing an AI-driven multimodal framework.
We proposed MultiST, a unified model that jointly learns from gene expression and histological images to derive maximally informed representations, enabling robust downstream analyses and biomarker discovery.

% As illustrated in Fig.~\ref{fig:framework}, MultiST jointly learns from molecular and morphological modalities to generate biologically meaningful spot-level embeddings that support a variety of downstream tasks, including spatial domain clustering, trajectory inference, and CCI analysis. The framework consists of three major components:
% 【(1) a Maximum-guided gene expression encoder that models both local expression features and spatial dependencies among neighboring spots;
% (2) a color-normalized image encoder based on a pretrained vision transformer that extracts consistent and biologically relevant morphological features; and
% (3) a cross-attention fusion module that integrates gene and image embeddings into a unified latent representation.】
As illustrated in Fig.~\ref{fig:framework}, our framework comprises three major modules.
(1) Input Data Module: We utilized Visium-based spatial transcriptomics data and paired Hematoxylin and Eosin (H\&E)–stained images as multimodal inputs, which jointly provide gene expression matrices, spatial coordinates, and morphological context from the same tissue.
(2) Spatially Guided Multimodal Integration Module: This module learns joint molecular-morphological representations via parallel encoding and cross-modal fusion.
Specifically, we first designed a graph-based spatial gene representation encoder to capture local expression patterns and spatial dependencies among neighboring spots, producing a latent representation. Then, we introduced a color-normalized image encoder based on a pretrained CLIP-ViT model to extract consistent and biologically meaningful morphological features.
We further aligned the latent representation and the image embedding using a cross-attention fusion mechanism and 
smoothed the embeddings via label diffusion to achieve spatial coherence across modalities.
(3) Downstream Analysis Module: Finally, we applied the combined embeddings to a variety of spatial transcriptomics tasks, including spatial domain identification, pseudotime trajectory inference, and CCI analysis.
In the following, we present the Spatially Guided Multimodal Integration Module in detail.

% These analyses enabled us to biologically interpret spatial organization, gene expression variation, and intercellular communication within complex tissue microenvironments.

% \vspace{-8pt}
\subsection{Graph-Based Spatial Gene Representation Learning}

\begin{figure}
    \centering
    \includegraphics[width=\linewidth]{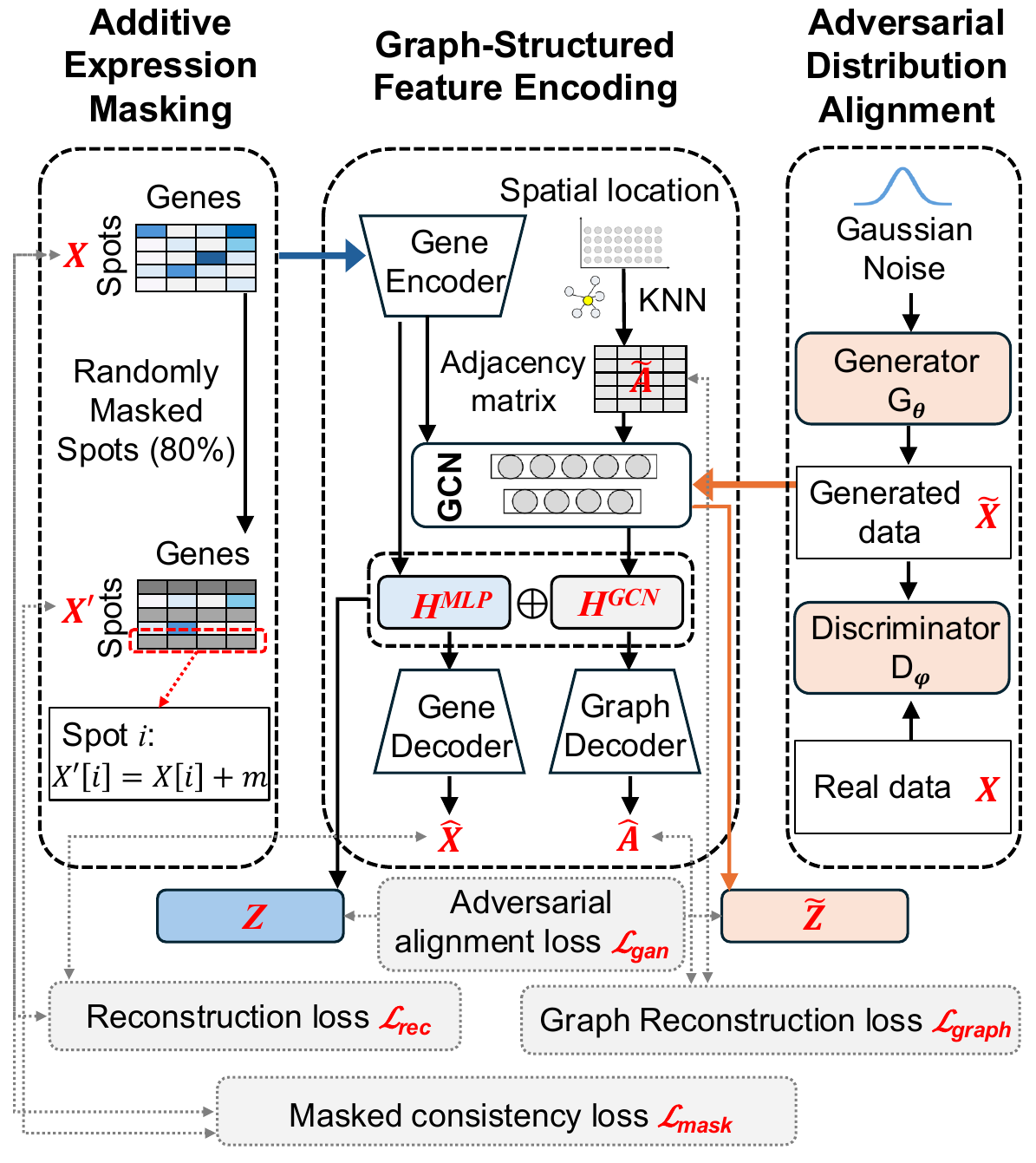}
    \vspace{-18pt}
    \caption{\textbf{Graph-based spatial gene representation learning framework.} The encoder integrates expression and spatial features into latent representations, with Fisher–MMD adversarial alignment and refined with deep embedded clustering.
    % The graph-based encoder integrates expression features and spatial structure into coherent latent representations, enhanced by Fisher–MMD–based adversarial alignment and refined via deep embedded clustering.
    }
    \label{fig:gene}
    \vspace{-15pt}
\end{figure}

Spatial transcriptomics data contain rich molecular and spatial information, where gene expression profiles are inherently influenced by the local tissue microenvironment. 
To preserve these spatial dependencies during representation learning, we designed a spatially informed gene encoder that integrates expression variation and spatial topology into a unified latent representation. 
This module consists of three major components: additive expression masking, graph-structured feature encoding and adversarial distribution alignment(Fig.~\ref{fig:gene}).
\subsubsection{Additive Expression Masking}
To enhance the robustness of feature learning under sparse or noisy
measurement conditions, we introduced an additive masking strategy
directly applied to the spatial expression vectors before graph
construction. Given the spatial gene expression matrix
$\mathbf{X}\in\mathbb{R}^{N\times G}$, where $N$ denotes the number of
spatial spots and $G$ represents the number of highly variable genes
(HVGs).
The HVGs were identified using the Seurat v3 
selection approach~\citep{seuratV3} implemented in Scanpy~\citep{scanpy},
and the top 2{,}000 genes were retained.
% identified using the Seurat v3 selection
% approach~\citep{seuratV3} implemented in Scanpy~\citep{scanpy}, retaining
% the top 2{,}000 HVGs. 
We randomly selected a subset of spots
$\mathcal{S}_{\text{mask}}\subset\{1,\dots,N\}$ (default ratio 80\%) for
masking.

For each masked spot $i\in\mathcal{S}_{\text{mask}}$, we injected a
learnable perturbation $\mathbf{m}\in\mathbb{R}^{G}$,
which is initialized as a trainable parameter and jointly updated
with the encoder during optimization.
The masked expression vectors $\mathbf{X}'$ was defined as:
\vspace{-0.5em}
\begin{equation}
\mathbf{X}'[i]=
\begin{cases}
\mathbf{X}[i]+\mathbf{m}, & i\in\mathcal{S}_{\text{mask}},\\[3pt]
\mathbf{X}[i], & \text{otherwise.}
\end{cases}
\end{equation}

% This vector is initialized as a trainable parameter and optimized jointly
% with the encoder, allowing the model to learn an informative masking
% direction in the gene-expression space. 
% The perturbed feature matrix$\mathbf{X}'$ is defined as:
% This additive perturbation maintains the global expression distribution
% while forcing the encoder to recover masked expression vectors from
% their spatial context, thereby improving representation robustness and
% spatial imputation capability.
\vspace{-0.5em}
This additive perturbation preserves the overall expression distribution while encouraging the encoder to reconstruct masked nodes from their spatial context, 
thus improving representation robustness and spatial imputation capability.
This module is supervised by a masked consistency loss
$\mathcal{L}_{\text{mask}}$, which enforces agreement between the
reconstructed and original expression profiles of masked spots.

% \subsubsection{Graph-Structured Feature Encoding}
%%%variable
\subsubsection{Graph-Structured Feature Encoding}

Based on the masked input matrix $\mathbf{X}' \in \mathbb{R}^{N \times G}$, we defined a spatial graph 
$\mathcal{G} = (\mathcal{V}, \mathcal{E}, \mathbf{X}')$ to model spatial dependencies among spots.  
 In the spatial graph $\mathcal{G}$, each spatial spot $i \in \{1, \ldots, N\}$ is represented by a node $v_i$, and the node set 
$\mathcal{V} = \{v_1, \ldots, v_N\}$ contains all spots. Each node $v_i$ is associated with a feature vector 
$\mathbf{x}_i \in \mathbb{R}^G$, corresponding to the $i$-th row of $\mathbf{X}'$, i.e., $\mathbf{x}_i = \mathbf{X}'[i]$.  
The edge set $\mathcal{E}$ is constructed by connecting each node to its $k$ nearest neighbors (KNN) 
in Euclidean space based on spot coordinates, resulting in an undirected edge $e_{ij} = (v_i, v_j) \in \mathcal{E}$ 
between spatially adjacent spots. 
The corresponding adjacency matrix $\mathbf{A} \in \{0,1\}^{N \times N}$ indicates the connectivity between nodes, where each entry $\mathbf{A}_{ij}$ is 1 if nodes $v_i$ and $v_j$ are connected by an edge, and 0 otherwise. To preserve each node’s own information,  we also added self-loops  to $\mathcal{E}$, i.e., $(v_i, v_i) \in \mathcal{E}$, and the adjacency matrix was symmetrically normalized as
\vspace{-0.4em}
\begin{equation}
\tilde{\mathbf{A}} = D^{-\frac{1}{2}}(\mathbf{A})D^{-\frac{1}{2}},
\vspace{-0.4em}
\end{equation}
where $D$ is the degree matrix of $\mathbf{A}$. %This normalization ensures numerical stability and balanced feature aggregation across nodes with varying degrees.

% The spatial graph $\mathcal{G}$ is represented by an adjacency matrix 
% $\mathbf{A} \in \{0,1\}^{N \times N}$, where $\mathbf{A}_{ij} = 1$ if $(v_i, v_j) \in \mathcal{E}$ 
% and $0$ otherwise. 
% To explicitly encode spatial structure, we construct $\mathbf{A}$ by connecting each node $v_i$ to its $k$ nearest neighbors under the Euclidean distance metric, resulting in a symmetric, unweighted adjacency matrix. 
% Self-loops are added to preserve each node’s self-information, and the adjacency matrix 
% is symmetrically normalized as
% \vspace{-0.5em}
% \begin{equation}
% \tilde{\mathbf{A}} = D^{-\frac{1}{2}}(\mathbf{A}+I)D^{-\frac{1}{2}},
% \vspace{-0.5em}
% \end{equation}
% where $D$ is the degree matrix of $\mathbf{A}+I$. 
% This normalization ensures numerical stability and balanced feature aggregation among nodes with varying degrees.

To jointly capture nonlinear expression features and spatial dependencies, the model utlized a dual-branch encoder with a multilayer perceptron (MLP) and a graph convolutional network (GCN). 
The MLP branch encoded each spot independently through two fully connected layers with batch normalization and ELU activation, producing local embeddings:
\vspace{-0.3em}
\begin{equation}
H^{\text{MLP}} = \text{MLP}(\mathbf{X}') \in \mathbb{R}^{N \times d_1},
\vspace{-0.4em}
\end{equation}
where $d_1$ denotes the latent dimension of the MLP branch.
These local embeddings were further processed by a two-layer GCN to
incorporate spatial neighborhood information. Using
$\mathbf{H}^{(0)} = \mathbf{H}^{\text{MLP}}$ as the initial node
representations, each GCN layer updated the features according to
\vspace{-0.5em}
\begin{equation}
  \mathbf{H}^{(l+1)} = \mathrm{ReLU}\!\left(
    \tilde{\mathbf{A}}\,\mathbf{H}^{(l)}\mathbf{W}^{(l)}
  \right),\quad l = 0,1,
  \vspace{-0.8em}
\end{equation}
where $H^{(l)}$ is the node representation at layer $l$, $\mathbf{W}^{(l)}$ is the trainable weight matrix of that layer, and $\text{ReLU}(\cdot)$ is the activation function. The final GCN output is denoted as \( H^{\text{GCN}} = H^{(2)} \in \mathbb{R}^{N \times d_2}\), where $d_2$ is the output dimension of the GCN branch.
% \( H^{\text{GCN}} = H^{(2)} \).
% The resulting Descripency informed representation was obtained by concatenating the two branches:
The resulting discrepancy-informed representation was obtained by concatenating the outputs of the two branches:
\vspace{-0.5em}
\begin{equation}
\mathbf{Z} = [H^{\text{MLP}} \| H^{\text{GCN}}] \in \mathbb{R}^{N \times d_z}, \quad d_z = d_1 + d_2.
\vspace{-0.4em}
\end{equation}
% where $d_2$ is the output dimension of the GCN branch and $d_z$ was the total latent dimensionality.

To maintain biological consistency and reconstruct masked information, the model integrated two decoding paths. 
The expression decoder reconstructed the original expression matrix from $\mathbf{Z}$ via a graph-based decoding layer:
\vspace{-0.5em}
\begin{equation}
\hat{\mathbf{X}} = \text{GCN}_{\text{dec}}(\mathbf{Z}, \tilde{\mathbf{A}}) \in \mathbb{R}^{N \times G},
\vspace{-0.5em}
\end{equation}
where $\hat{\mathbf{X}}$ denotes the reconstructed expression profiles. 
Let $\mathbf{z}_i, \mathbf{z}_j \in \mathbb{R}^d_z$ denote the $d$-dimensional latent representations 
of nodes $v_i$ and $v_j$, respectively. 
Meanwhile, the graph decoder predicted the edge probability between two nodes based on latent similarity:
\vspace{-0.8em}
\begin{equation}
\hat{A}_{ij} = Sigmoid(z_i^\top z_j),
\vspace{-0.5em}
\end{equation}
% This dual-path decoding designed jointly reconstructs gene expression and spatial adjacency, ensuring biologically consistent and Descripency coherent representations.
This dual-path decoding jointly reconstructed gene expression and spatial adjacency, ensuring biologically consistent and discrepancy-informed representations.

This dual-path decoding corresponds to two complementary training objectives:
the expression reconstruction loss $\mathcal{L}_{\text{rec}}$, which enforces accurate recovery of gene expression profiles,
and the graph topology loss $\mathcal{L}_{\text{graph}}$, which preserves spatial adjacency in the latent space.

\subsubsection{Adversarial Distribution
Alignment }

To achieve a biologically consistent and continuous latent representation $\mathbf{Z}$, 
we introduced an adversarial distribution alignment via a Generative Adversarial Network (GAN)~\citep{GAN}.
Let $G_{\theta} : \mathbb{R}^{N \times G} \to \mathbb{R}^{N \times G}$ 
and $D_{\phi} : \mathbb{R}^{N \times G} \to \mathbb{R}$ 
denote the generator and discriminator.
% be a generator and discriminator, respectively. 
Given noise $\mathbf{n} \sim \mathcal{N}(\mathbf{0}, I)$,
the generator produced synthetic samples 
$\tilde{\mathbf{X}}_{\text{gen}} = G_{\theta}(\mathbf{n})$.
% $\hat{\mathbf{X}}_{\mathrm{gen}} = G_{\theta}(\mathbf{n})$
 Under the standard minimax objective
$
\min_{\theta} \max_{\phi} 
\mathcal{L}_{\mathrm{GAN}}(\theta,\phi),
$
the generator was trained such that the induced distribution~$q$ of
generated samples approaches the empirical data distribution~$p$.
% As a result, 
% Consequently, the encoder learns latent embeddings 
% ~$\mathbf{Z}$ aligned with real gene expression geometry.
As a result, the encoder  learns latent embeddings~$\mathbf{Z}$ whose induced manifold
aligns with the geometry of real gene-expression data.

% drives distribution of generated sample  toward true data distribution.
% The encoder thus learns latent embeddings $\mathbf{Z}$ whose induced manifold matches the real gene-expression geometry.

To further enhance higher-order distributional alignment, 
we proposed a Fisher-kernel-induced Maximum Mean Discrepancy (Fisher-MMD)~\citep{MMD}.
% Let $p$ denote the empirical distribution of real gene-expression vectors,
% and $q$ the distribution generated by $G_{\theta}$. 
Let $k_{F}$ denote the Fisher kernel~\citep{fisher} associated with the
generator parameterization. 
The Fisher-MMD between $p$ and $q$ was defined as:

% {\vspace{-13pt}
{\vspace{-10pt}
\fontsize{9.5pt}{9.5pt}\selectfont
% \begin{equation}
% \begin{split}
% \mathrm{MMD}^{2}_{k_F}(p,q)
% = &
% \ \mathbb{E}_{u,\,u' \sim p}\!\left[k_F(u,u')\right] 
%  +\mathbb{E}_{g,\,g' \sim q}\!\left[k_F(g,g')\right] \\
% & -2\,\mathbb{E}_{u \sim p,\, g \sim q}\!\left[k_F(u,g)\right],
% \end{split}
% \vspace{-1.1em}
% \end{equation}
\begin{equation}
\begin{aligned}
\mathrm{MMD}^2_{k_F}(p,q)
&= \mathbb{E}_{u,u' \sim p}[k_F(u,u')] 
 + \mathbb{E}_{g,g' \sim q}[k_F(g,g')] \\
&\quad - 2\,\mathbb{E}_{u \sim p,\, g \sim q}[k_F(u,g)] ,
\end{aligned}
\vspace{-2pt}
% \vspace{-0.4em}
\end{equation}
% \vspace{-0.3em}
}
\noindent\hspace{-0.4em}where $u,u'$ are independent samples from $p$ and 
$g,g'$ are independent samples from $q$.

% where $u,u'$ are independent samples from $p$ and 
% $g,g'$ are independent samples from $q$.

The generative regularization loss 
$\mathcal{L}_{\mathrm{gan}}$ motivates $\mathbf{Z}$ to form a smooth, biologically meaningful manifold
where real and generated samples share consistent higher-order structure.

% \vspace{-15pt}
\subsection{Color-Normalized Image Feature Extraction}

\begin{figure}
    \centering
    \includegraphics[width=\linewidth]{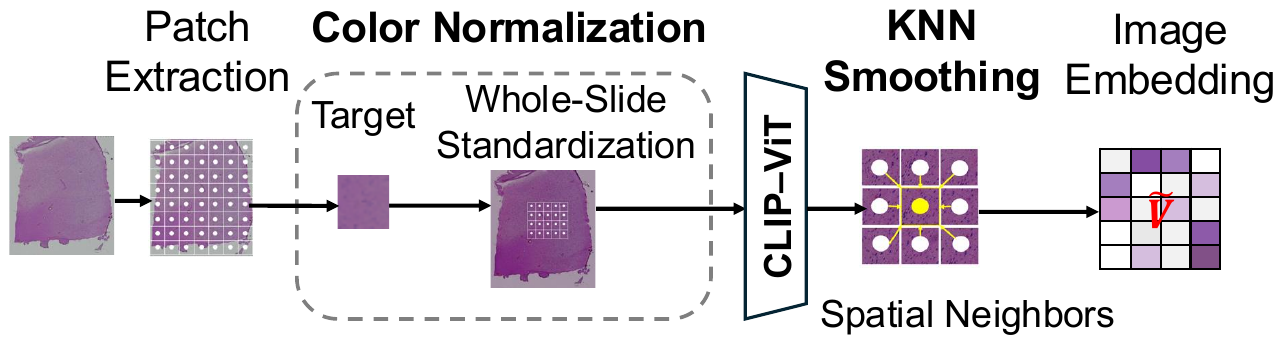}
    \vspace{-18pt}
    \caption{
    \textbf{Color-Normalized Image Feature Extraction pipeline.}
    Coordinate-aligned patches are encoded with a pretrained CLIP–ViT model, followed by KNN smoothing to enhance local consistency of the embeddings ($\tilde{V}$).%Coordinate-aligned patches are color-normalized using representative targets and encoded by a pretrained CLIP–ViT model.
% KNN-based smoothing enhances local consistency of the resulting image embeddings ($\tilde{V}$).
}
    \label{fig:HEimage}
    \vspace{-15pt}
\end{figure}

% To extract consistent and biologically meaningful morphological features from H\&E-stained tissue images, we adopted a three-stage pipeline (Fig.~\ref{fig:HEimage}): (1) coordinate-aligned patch extraction, (2) color-normalized target selection and whole-slide standardization, and (3) patch-level visual encoding with spatial smoothing.

% \subsection{Color-Normalized Image Feature Extraction}
To extract biologically meaningful morphological features from H\&E-stained tissue sections, we developed a three-step color-normalized image feature extraction pipeline (Fig.~\ref{fig:HEimage}).
The pipeline extracts coordinate-aligned patches, performs reference-based color normalization, and encodes patch-level features with spatial smoothing using a CLIP image encoder.
Implementation and mathematical details are provided in Appendix~A.2 (Supplementary Materials).
% To extract consistent and biologically meaningful morphological features from H\&E-stained tissue sections, we developed a three-step color-normalized image feature extraction pipeline (Fig.~\ref{fig:HEimage}). 
% The pipeline performs coordinate-aligned patch extraction, applies color normalization via reference patch selection, 
% and encodes patch-level visual features with spatial smoothing using a CLIP image encoder. 
% Implementation and mathematical details are provided in Appendix~A.2 (Supplementary Materials).
% Full implementation details and mathematical formulations are provided in Appendix~A.2 (Supplementary Materials).

% we developed a three-step color-normalized image feature extraction pipeline (Fig.~\ref{fig:HEimage}): 
% (1) coordinate-aligned patch extraction, 
% (2) stain normalization using a curated reference set of high-quality patches, and 
% (3) CLIP-based patch-level visual encoding with spatial smoothing. 
% Full implementation details and mathematical formulations are provided in Appendix~A.

\subsubsection{Coordinate-Aligned Patch Extraction}
Let $\mathbf{S} = \{\mathbf{s}_i\}_{i=1}^{N} \in \mathbb{R}^{N \times 2}$ denote the spatial coordinates of $N$ spots, 
where $\mathbf{s}_i = (x_i, y_i)$. 
To align the spatial domain with the histology image resolution, the coordinates were scaled by a predefined scale factor $\gamma$ obtained from the Visium metadata. Specifically, $\gamma$ corresponds to the \texttt{tissue\_hires\_scalef} parameter. 
For each spot, we extracted a square Red–Green–Blue (RGB) patch of size $64 \times 64$ centered at $(\gamma x_i, \gamma y_i)$. 
Coordinates were used directly with appropriate padding to ensure all patches were fully contained within the image. For spots located near the image boundary, reflection padding was applied to preserve contextual information without introducing artificial edges.

\subsubsection{Color Normalization via Target Patch Selection}
To mitigate inter-slide staining variability, we adopted a two-step normalization strategy. 
First, high-quality tissue patches were identified using a composite visual scoring function that considers tissue coverage, contrast, texture, and color diversity. 
A diverse target set was then selected through $K$-means clustering, and their foreground pixel statistics were used to derive global target mean--variance parameters.

Second, whole-slide stain appearance was harmonized via a channel-wise affine transformation to the raw RGB image.  
Let $\mu_{\text{raw}}$ and $\sigma_{\text{raw}}$ denote the mean and standard deviation of foreground pixels in the original slide $\mathbf{I}_{\text{raw}}$,  
and let $\mu_{\text{tgt}}$ and $\sigma_{\text{tgt}}$ 
aggregated target statistics from selected target patches.
% represent the aggregated target statistics obtained from the selected reference patches.  
The normalized image $\mathbf{I}_{\text{norm}}$ was computed as:
\vspace{-0.3em}
\begin{equation}
\label{eq:stain_norm}
\mathbf{I}_{\text{norm}}
=
\frac{\mathbf{I}_{\text{raw}} - \mu_{\text{raw}}}{\sigma_{\text{raw}}}
\cdot \sigma_{\text{tgt}}
+ \mu_{\text{tgt}}.
\vspace{-0.4em}
\end{equation}
This linear transformation preserves local morphology while standardizing color distribution across slides.  
% Additional implementation details of patch scoring, diversity selection, and foreground estimation are provided in Appendix~A.2--A.3.

\subsubsection{Patch-Level Visual Encoding and Spatial Smoothing}
After color normalization, $64\times 64$ coordinate-aligned tissue patches were encoded using a pretrained CLIP--ViT model~\citep{clip}, yielding patch-level visual embeddings $\mathbf{v}_i$ for each spot $i$.

To enhance spatial coherence and suppress local noise, we applied a Gaussian-weighted KNN smoothing over neighboring spots.  
For each spot $i$, let $\mathcal{N}(i)$ denote its spatial neighbors and $d_{ij}$ the Euclidean distance between spots $i$ and $j$.  
The smoothed feature was computed as:
\vspace{-1.1em}
% \begin{equation}
% \tilde{\mathbf{v}}_i = (1-\lambda)\,\mathbf{v}_i
% + \lambda \sum_{j \in \mathcal{N}(i)} w_{ij}\,\mathbf{v}_j,
% \qquad
% w_{ij} = \frac{\exp(-d_{ij}^2 / 2\sigma^2)}{C_i},
% \end{equation}
{

\fontsize{9.5pt}{9.5pt}\selectfont
\begin{equation}
\tilde{\mathbf{v}}_i = (1-\lambda)\mathbf{v}_i
+ \lambda \sum_{j\in\mathcal{N}(i)} w_{ij}\mathbf{v}_j, \quad
w_{ij} = \frac{\exp(-d_{ij}^2/2\sigma^2)}{C_i},
\vspace{-0.4em}
\end{equation}
}
\noindent\hspace{-0.4em}where $C_i$ normalizes the weights so that $\sum_j w_{ij}=1$.  
The set $\tilde{\mathbf{V}} = \{\tilde{\mathbf{v}}_i\}_{i=1}^N$ represents the learned image embeddings, combined with the latent representations through multimodal fusion. This smoothing step reduces spurious visual variability while preserving fine-grained morphological structure.

% The resulting feature set $\tilde{\mathbf{V}}=\{\tilde{\mathbf{v}}_i\}_{i=1}^N$ was used as the visual input for multimodal fusion.

% This smoothing step reduces spurious visual variability while preserving fine-grained morphological structure.

\subsection{Multi-modal Spatial Clustering}

\subsubsection{Deep Embedded Clustering Alignment}

Upon obtaining stable latent features $\mathbf{Z}$,  
we performed deep embedded clustering (DEC)~\citep{DEC} to identify spatial domains that share similar molecular and spatial characteristics. 
Given $\mathbf{Z} = [\mathbf{z}_i]_{i=1}^N \in \mathbb{R}^{N \times d}$,  
cluster centroids $\{\boldsymbol{\mu}_j\}_{j=1}^{K}$ were initialized using KMeans:
$\boldsymbol{\mu}_j^{(0)} \leftarrow \mathrm{KMeans}(\mathbf{Z}, K)$.

% For $\alpha > 0$, the Student–$t$ kernel yields the soft assignment
% \begin{equation}
% \label{eq:dec_q_side}
% \begin{aligned}
% q_{ij}
% &=
% \frac{
% \left( 1 + \frac{\|\mathbf{z}_i - \boldsymbol{\mu}_j\|^{2}}{\alpha} \right)^{-\beta}
% }{
% \sum_{j'=1}^{K}
% \left( 1 + \frac{\|\mathbf{z}_i - \boldsymbol{\mu}_{j'}\|^{2}}{\alpha} \right)^{-\beta}
% } ,
% \end{aligned}
% \hspace{1em}  
% \beta = \frac{\alpha + 1}{2},
% \hspace{1em}  
% \alpha = 1.
% \end{equation}

For $\alpha > 0$, the Student--$t$ kernel yields the soft assignment
\vspace{-0.4em}
\begin{equation}
\label{eq:dec_q}
q_{ij}
=
\frac{
\left( 1 + \frac{\lVert \mathbf{z}_i - \boldsymbol{\mu}_j \rVert^2}{\alpha} \right)^{-\beta}
}{
\sum_{j'=1}^{K}
\left( 1 + \frac{\lVert \mathbf{z}_i - \boldsymbol{\mu}_{j'} \rVert^2}{\alpha} \right)^{-\beta}
},
\quad
\beta = \frac{\alpha+1}{2}.
\vspace{-0.4em}
\end{equation}
In practice, we followed the standard choice $\alpha = 1$.
This yields the soft assignment matrix 
$\mathbf{Q} = [q_{ij}] \in \mathbb{R}^{N \times K}$.  
The target distribution was defined as
\vspace{-0.8em}
\begin{equation}
p_{ij}
= \frac{
q_{ij}^{2} / \sum_{i=1}^{N} q_{ij}
}{
\sum_{j'=1}^{K} q_{ij'}^{2} / \sum_{i=1}^{N} q_{ij'}
},
\vspace{-0.4em}
\end{equation}
enhances confident assignments while normalizing clusters.
% amplifies confident assignments while normalizing cluster contributions.

DEC minimizes the KL divergence
\vspace{-0.4em}
\begin{equation}
\mathcal{L}_{\mathrm{DEC}}
= \sum_{i=1}^{N} \sum_{j=1}^{K} p_{ij} \log \frac{p_{ij}}{q_{ij}},
\vspace{-0.4em}
\end{equation}
% jointly refining both centroids and encoder parameters.  
% Combined with Fisher–MMD regularization on $\mathbf{Z}$,  
% the latent space evolves toward spatially and molecularly coherent domains.
jointly updating $\{\boldsymbol{\mu}_j\}$ and the encoder parameters (hence $\mathbf{Z}$).
Combined with the Fisher–MMD regularization on $\mathbf{Z}$, 
latent domains are molecularly and spatially coherent.

\subsubsection{Cross-Attention–based Multimodal Fusion}

To effectively integrate Fisher-MMD aligned gene expression and histological image features, 
we employed a bidirectional cross-attention mechanism that enables joint and localized interaction 
between molecular and visual representations. 
This design allows each modality to adaptively attend to complementary information from the other, 
achieving multimodal alignment.

Let $\mathbf{z}_i \in \mathbb{R}^{d_z}$ and $\tilde{\mathbf{v}}_i \in \mathbb{R}^{d_v}$ denote the expression-derived embedding obtained from the graph-based encoder and the corresponding smoothed image patch feature of spot $i$, respectively.
Both modalities were first projected into a shared latent space using learnable linear layers:
\vspace{-0.1em}
\begin{equation}
\mathbf{Q}^{(G)}_i = W_Q^{(G)} \mathbf{z}_i,  \;
\mathbf{K}^{(I)}_i = W_K^{(I)} \tilde{\mathbf{v}}_i,  \;
\mathbf{V}^{(I)}_i = W_V^{(I)} \tilde{\mathbf{v}}_i,
\vspace{-0.5em}
\end{equation}
\vspace{-0.5em}
\begin{equation}
\mathbf{Q}^{(I)}_i = W_Q^{(I)} \tilde{\mathbf{v}}_i,  \;
\mathbf{K}^{(G)}_i = W_K^{(G)} \mathbf{z}_i,  \;
\mathbf{V}^{(G)}_i = W_V^{(G)} \mathbf{z}_i,
\vspace{-0.3em}
\end{equation}
where $W_Q^{(G)}, W_K^{(G)}, W_V^{(G)} \in \mathbb{R}^{d \times d_z}$ and 
$W_Q^{(I)}, W_K^{(I)}, W_V^{(I)} \in \mathbb{R}^{d \times d_v}$ 
project the gene and image embeddings into a shared attention space of dimension $d$.

% where $W_Q^{(\cdot)}, W_K^{(\cdot)}, W_V^{(\cdot)}$ were modality-specific projection matrices,
% and $d$ was the shared attention dimension.

%  where $W_Q \in \mathbb{R}^{d \times d_z}$ and $W_K, W_V \in \mathbb{R}^{d \times d_v}$ 
% % are modality-specific projection matrices, and $d$ denotes the shared attention dimension.

% The bidirectional fusion was realized by performing cross-attention in both directions:
The bidirectional fusion was realized by performing multi-head cross-attention in both directions:
\vspace{-0.2em}
\begin{equation}
\begin{aligned}
\mathbf{h}_i^{G } &= 
\text{softmax}\!\left(\frac{\mathbf{Q}^{(G)}_i (\mathbf{K}^{(I)}_i)^\top}{\sqrt{d}}\right)
\mathbf{V}^{(I)}_i + \mathbf{z}_i, \\
\mathbf{h}_i^{I } &=
\text{softmax}\!\left(\frac{\mathbf{Q}^{(I)}_i (\mathbf{K}^{(G)}_i)^\top}{\sqrt{d}}\right)
\mathbf{V}^{(G)}_i + \tilde{\mathbf{v}}_i.
\end{aligned}
\vspace{-0.5em}
\end{equation}

The final fused representation was obtained by a weighted combination of the two enhanced embeddings:
\vspace{-0.4em}
\begin{equation}
\mathbf{h}_i^{\text{fusion}} = 
\alpha \mathbf{h}_i^{G } + (1 - \alpha) \mathbf{h}_i^{I },
\vspace{-0.4em}
\end{equation}
where $\alpha \in [0, 1]$ controles the relative contribution of gene- and image-enhanced features was set to $0.7$ in all experiments.
% The fused embeddings $\{\mathbf{h}_i^{\text{fusion}}\}_{i=1}^{N}$ are used for downstream tasks.

For datasets with clear tissue organization and stable spatial patterns (e.g., human dorsolateral prefrontal cortex), 
we adopted the unidirectional variant (Image $\rightarrow$ Gene) for improved stability and interpretability, 
while retaining the same fusion framework and parameterization.

\subsubsection{Multi-modal Clustering and Spatial Label Diffusion}

To identify MMD coherent tissue domains, we performed unsupervised clustering on the fused multimodal latent representations, followed by spatial label propagation to enforce local consistency. 
% The complete procedure was summarized in Algorithm~S1 (Supplementary Materials).
Specifically, let $\mathbf{H}^{\text{fusion}} = [\mathbf{h}_1^{\text{fusion}}, \dots, \mathbf{h}_N^{\text{fusion}}]^\top \in \mathbb{R}^{N \times d_h}$ 
denote the multimodal embeddings obtained from the cross-attention fusion module, 
where each $\mathbf{h}_i^{\text{fusion}}$ represents the image-informed gene expression embedding of spot $i$. 
We applied a Gaussian mixture model (GMM) with Mclust~\citep{mclust}  to partition $\mathbf{H}^{\text{fusion}}$ into $C$ spatial clusters.
Mclust modeled the multimodal features $\mathbf{H}^{\text{fusion}}$ as a mixture of Gaussians and estimates posterior cluster assignment probabilities. 

To enhance spatial smoothness, we further introduced a label refinement module based on spatial label propagation. Let $\mathcal{X} = \{x_i\}_{i=1}^N$ be the spatial coordinates and $\mathcal{L} = \{\ell_i\}_{i=1}^N$ the initial cluster labels. We constructed a KNN spatial graph and defined the adjacency matrix $W$ using a Gaussian kernel:
\vspace{-0.3em}
\begin{equation}
W_{ij} = 
\begin{cases}
\exp\left(-\frac{\|x_i - x_j\|^2}{\sigma^2}\right), & \text{if } x_j \in \mathcal{N}_k(x_i), \\
0, & \text{otherwise},
\end{cases}
\vspace{-0.4em}
\end{equation}
where $\sigma$ is set to the median of all pairwise distances in the neighborhood.

% To ensure stability, we identified high-confidence anchor spots within each cluster based on intra-cluster agreement scores:
To ensure stability, we identified high-confidence anchor spots in each cluster using intra-cluster agreement scores:
\vspace{-0.3em}
\begin{equation}
\text{Agree}_i = \sum_{j} W_{ij} \cdot \mathbb{I}[\ell_j = \ell_i],
\vspace{-0.5em}
\end{equation}
and retain the top 1\% most consistent spots per cluster as anchors.

Let $Y \in \mathbb{R}^{N \times C}$ denote the initial one-hot label matrix. Label propagation was performed via the following iterative update:
\vspace{-0.3em}
\begin{equation}
Y^{(t+1)} = \text{Normalize}(W Y^{(t)}),
\vspace{-0.5em}
\end{equation}
with anchor rows held fixed across iterations. Final labels were obtained by:
\vspace{-0.3em}
\begin{equation}
\hat{\ell}_i = \arg\max_j Y_{ij}.
\vspace{-0.5em}
\end{equation}

This two-stage clustering pipeline, comprising probabilistic clustering in fused feature space and MMD aware label refinement, balances feature discriminability and spatial continuity, enabling robust delineation of heterogeneous tissue structures.

\subsection{Training Objectives}
The training objective of the proposed framework integrated multi-level losses across three stages. Each stage was designed to progressively enhance representation quality, ranging from feature reconstruction and structural preservation to clustering consistency and generative regularization, thereby ensuring stable and biologically meaningful multimodal embeddings.

% \textbf{Stage I: Structure-Preserving Pretraining.}  
\subsubsection{Stage I: Gene Embedding Learning}
In the pretraining stage, the model aimed to  learn latent representations that preserve both gene expression features and spatial topology through an integrated autoencoder–graph convolutional framework.  
The overall pretraining objective was formulated as:
\vspace{-0.4em}
\begin{equation}
    \mathcal{L}_{\text{gene}} =
\lambda_{\text{rec}}\mathcal{L}_{\text{rec}} +
\lambda_{\text{graph}}\mathcal{L}_{\text{graph}} +
\lambda_{\text{mask}}\mathcal{L}_{\text{mask}} +
\lambda_{\text{gan}}\mathcal{L}_{\text{gan}}.
\vspace{-0.3em}
\end{equation}

Here, $\mathcal{L}_{\text{rec}} = \|\hat{\mathbf{X}} - \mathbf{X}\|_F^2$ ensured accurate reconstruction of gene expression profiles.  
$\mathcal{L}_{\text{graph}}$ preserved spatial topology by enforcing binary cross-entropy between the predicted and true adjacency matrices, together with a KL divergence term that regularizes the latent embeddings.
% $\mathcal{L}_{\text{graph}}$ preserved spatial topology by enforcing binary cross-entropy between the predicted and true adjacency matrices, together with an $L_2$ regularization term on the latent representations.

To enhance robustness to noise and missing input, a masked consistency
 loss  was introduced to encourage masked nodes to retain consistent representations after reconstruction:
\vspace{-0.5em}
\begin{equation}
    \mathcal{L}_{\text{mask}} =
\mathbb{E}_{i}\left[\big(1 - \langle \hat{\mathbf{x}}_i , \mathbf{x}_i \rangle\big)^{\alpha}\right],
\vspace{-0.4em}
\end{equation}
% where $\hat{\mathbf{x}}_i$ and $\mathbf{x}_i$ denoted the reconstructed and original features of the $i$-th spot, and $\alpha = 3$, which controls the similarity penalty strength. 
where $\hat{x}_i$ and $x_i$ denote the reconstructed and original 
features of the $i$-th spot. The exponent $\alpha$ controls the 
sharpness of the similarity penalty and is set to $3$ in our 
experiments.

\vspace{0.1em}
To smooth the latent space and improve generalization,  
a Fisher-kernel-based Maximum Mean Discrepancy (Fisher–MMD) regularization was introduced.  
The generator $G_{\theta}$ produced synthetic samples $\tilde{\mathbf{X}}$,  
which were encoded into latent representations $\tilde{\mathbf{Z}} = f_{\text{enc}}(\tilde{\mathbf{X}})$.  
The distributional discrepancy between $\mathbf{Z}$ and $\tilde{\mathbf{Z}}$ was measured using the Fisher–MMD:
% \begin{equation}
%     \mathcal{L}_{\text{gan}} =
% \text{MMD}_{\text{Fisher}}(\mathbf{Z}, \tilde{\mathbf{Z}}),
% \quad
% K_{ij} = \langle \nabla_{\theta}G_{\theta}(x_i), \nabla_{\theta}G_{\theta}(y_j)\rangle,
% \end{equation}
\vspace{-0.6em}
\begin{equation}
\begin{aligned}
\mathcal{L}_{\text{gan}} &=
\text{MMD}_{\text{Fisher}}(\mathbf{Z}, \tilde{\mathbf{Z}}), \\
K_{ij} &=
\langle \nabla_{\theta}G_{\theta}(x_i),
\nabla_{\theta}G_{\theta}(y_j)\rangle ,
\end{aligned}
\vspace{-0.4em}
\end{equation}

% {
% \fontsize{9.5pt}{9.5pt}\selectfont
% \begin{equation}
%     \mathcal{L}_{\text{gan}} =
% \text{MMD}_{\text{Fisher}}(\mathbf{Z}, \tilde{\mathbf{Z}}),
% \;
% K_{ij} = \langle \nabla_{\theta}G_{\theta}(x_i), \nabla_{\theta}G_{\theta}(y_j)\rangle,
% % \vspace{-0.3em}
% \end{equation}
% }
\noindent\hspace{-0.4em}where the Fisher kernel $K_{ij}$ quantifies the similarity between samples in the generator’s parameter space.

% \textbf{Stage II: Clustering Optimization and Generative Consistency.}  
\subsubsection{Stage II: Deep Embedded Clustering}
After pretraining, the model performed DEC and applied generative consistency constraint on the fused multimodal embeddings $\mathbf{H}^{fusion}$.  
The overall clustering objective was defined as:
\vspace{-0.3em}
\begin{equation}
    \mathcal{L}_{\text{dec}} =
\lambda_{\text{rec}}\mathcal{L}_{\text{rec}} +
\lambda_{\text{graph}}\mathcal{L}_{\text{graph}} +
\lambda_{\text{DEC}}\mathcal{L}_{\text{DEC}} +
\lambda_{\text{gan}}\mathcal{L}_{\text{cons}}.
\vspace{-0.3em}
\end{equation}
Here, $\mathcal{L}_{\text{rec}}$ and $\mathcal{L}_{\text{graph}}$ followed the same definitions as in Stage~I,  
ensuring accurate expression reconstruction and spatial structural consistency.  
$\mathcal{L}_{\text{DEC}}$ represented the KL-divergence objective based on the Student’s $t$-distribution (see Eq.~(8)), minimizing the divergence between the current soft assignment $Q$ and the target distribution $P$ to enhance cluster separability.

To further stabilize clustering boundaries, a generative consistency constraint was incorporated.  
As described earlier, the generator $G_{\theta}$ produced synthetic samples $\tilde{\mathbf{X}}$  
and their corresponding soft cluster assignments $\tilde{Q}$, defined as:
\vspace{-0.5em}
\begin{equation}
  \tilde{Q}_{ij} =
\frac{(1 + \alpha\|\tilde{\mathbf{z}}_i - \boldsymbol{\mu}_j\|^2)^{-\frac{\alpha+1}{2}}}
{\sum_{j'}(1 + \alpha\|\tilde{\mathbf{z}}_i - \boldsymbol{\mu}_{j'}\|^2)^{-\frac{\alpha+1}{2}}}.
\vspace{-0.5em}
\end{equation}

The model minimized the divergence between the real and generated clustering distributions,  
while aligning their latent representations via Fisher–MMD:
\vspace{-0.5em}
\begin{equation}
\mathcal{L}_{\text{cons}} =
\text{KL}(\tilde{Q} \parallel Q) +
\text{MMD}_{\text{Fisher}}(\mathbf{Z}, \tilde{\mathbf{Z}}).
\vspace{-0.5em}
\end{equation}
This consistency constraint enforces stable and coherent clustering under generative perturbations,  
enhancing the reliability of spatial domain delineation.

\subsubsection{Stage III: Cross-Attention Optimization.}  
The representations refined by the cross-attention module are further optimized through a combination of similarity distribution matching and cross-modal contrastive objectives, 
ensuring structural coherence and semantic correspondence across modalities.  
The overall loss function was defined as:
\vspace{-0.5em}
\begin{equation}
\mathcal{L}_{\text{cross}} =
\lambda_{\text{sdm}}\mathcal{L}_{\text{SDM}} +
\lambda_{\text{con}}\mathcal{L}_{\text{con}} +
\lambda_{\text{reg}}\mathcal{L}_{\text{reg}}.
\vspace{-0.4em}
\end{equation}

\textit{(1) Similarity Distribution Matching (SDM).}  
Let $\mathbf{H}^{(I)} = [\mathbf{h}_1^{(I)}, \mathbf{h}_2^{(I)}, \dots, \mathbf{h}_N^{(I)}]^\top$ 
and $\mathbf{H}^{(G)} = [\mathbf{h}_1^{(G)}, \mathbf{h}_2^{(G)}, \dots, \mathbf{h}_N^{(G)}]^\top$ 
denote denote the cross-attended embeddings from image and gene modalities.
% the matrices of cross-attended embeddings from the image and gene modalities, respectively. 
The SDM loss enforced structural consistency between the pairwise similarity distributions of the cross-attended embeddings 
across modalities:
% from the two modalities:
\vspace{-0.5em}
\begin{equation}
\mathcal{L}_{\text{SDM}} =
\frac{1}{2}
\big[
D_{\text{KL}}(\mathbf{P}^{(I)} \,\|\, \mathbf{P}^{(G)}) +
D_{\text{KL}}(\mathbf{P}^{(G)} \,\|\, \mathbf{P}^{(I)})
\big],    
\vspace{-0.5em}
\end{equation}
% where $\mathbf{P}^{(I)} = \text{softmax}(\mathbf{S}^{(I)}/\tau)$,  
% and $\mathbf{S}^{(I)}_{ij} = \langle \mathbf{h}_i^{(I)}, \mathbf{h}_j^{(I)} \rangle$  
% denoted the cosine similarity matrix computed from the attended image embeddings.  
where 
$S^{(I)}_{ij} = \langle h^{(I)}_i, h^{(I)}_j \rangle$ and 
$S^{(G)}_{ij} = \langle h^{(G)}_i, h^{(G)}_j \rangle$
denote the pairwise cosine similarities between the $i$-th and $j$-th spots in the image and gene modalities, respectively.
Here, $\tau$ is a temperature parameter that controls the sharpness of the similarity distribution.
In our experiments, we set $\tau = 0.12$.
This objective aligns the intra-modal similarity structures 
so that expression-derived and image-derived representations 
preserve consistent neighborhood relationships in the shared latent space.

\textit{(2) Cross-Modal Contrastive Alignment.}  
To further align paired representations across modalities,  
we adopted a symmetric cross-modal contrastive objective:
\vspace{-0.5em}
\begin{equation}
\begin{aligned}
\mathcal{L}_{\text{con}} =
\frac{1}{2N}
\sum_{i=1}^{N}
\Bigg[
&-\log
\frac{\exp(\langle \mathbf{h}_i^{(I)}, \mathbf{h}_i^{(G)} \rangle / \tau)}
{\sum_{j=1}^{N} \exp(\langle \mathbf{h}_i^{(I)}, \mathbf{h}_j^{(G)} \rangle / \tau)} \\
&-\log
\frac{\exp(\langle \mathbf{h}_i^{(G)}, \mathbf{h}_i^{(I)} \rangle / \tau)}
{\sum_{j=1}^{N} \exp(\langle \mathbf{h}_i^{(G)}, \mathbf{h}_j^{(I)} \rangle / \tau)}
\Bigg].
\end{aligned}
\vspace{-0.3em}
\end{equation}
This loss maximizes the similarity between matched image–gene pairs 
while contrasting them against mismatched ones, 
thus encouraging cross-modal semantic alignment.

\textit{(3) Regularization.}  
% To stabilize training and prevent modality imbalance,  
% we applied an $L_2$ regularization to the fused representations:
To stabilize training and balance modalities,
We applied L2 regularization on the fused representations
\vspace{-0.9em}
\begin{equation}
\mathcal{L}_{\text{reg}} =
\| \mathbf{H}^{(I)} \|_2 + 
\| \mathbf{H}^{(G)} \|_2,
\vspace{-0.5em}
\end{equation}
where $\mathbf{H}^{(I)}$ and $\mathbf{H}^{(G)}$ are defined as above.
This term constrained feature magnitude and prevents overfitting.

% To stabilize training and prevent modality imbalance,  
% we applied a lightweight regularization to the fused representations. Let $\mathbf{H}^{(I)} = [\mathbf{h}_1^{(I)}, \mathbf{h}_2^{(I)}, \dots, \mathbf{h}_N^{(I)}]^\top$ 
% and $\mathbf{H}^{(G)} = [\mathbf{h}_1^{(G)}, \mathbf{h}_2^{(G)}, \dots, \mathbf{h}_N^{(G)}]^\top$ 
% denote the matrices of cross-attended embeddings from the image and gene modalities, respectively. We define:
% \begin{equation}
% \mathcal{L}_{\text{reg}} =
% \| \mathbf{H}^{(I)} \|_2 + 
% \| \mathbf{H}^{(G)} \|_2,
% \end{equation}
% where $\mathbf{H}^{(I)}$ and $\mathbf{H}^{(G)}$ denoted 
% the matrices of cross-attended features from image and gene modalities, respectively.  
% This term constrained feature magnitude and prevents overfitting.

Together, these objectives ensure the fused embeddings are 
 semantically aligned and structurally coherent, 
resulting in robust and biologically interpretable multimodal representations.

\section{Datasets and Preprocessing}
We utilized spatial transcriptomic data generated with the 10× Genomics Visium platform, which captures gene expression across 55\,\textmu m-diameter barcoded spots. Each spot aggregates transcripts from multiple adjacent cells, enabling tissue-scale molecular profiling. In this study, we analyzed a total of 13 spatial transcriptomics sections derived from two human organs: the brain and the breast.
% Two representative spatial transcriptomics dataset types were analyzed in this study.

The human dorsolateral prefrontal cortex (DLPFC) dataset~\citep{dlpfc_1} consists of twelve sections from three healthy donors. Each section contains approximately 3,500--4,800 spots and has been manually annotated into six cortical layers and white matter, serving as a reference for evaluation. The human breast cancer (BRCA) dataset~\citep{SEDR} comprises a single tumor section with 3,798 spots, encompassing both malignant epithelial and immune-infiltrated regions, and thus supports downstream investigation of tumor heterogeneity and spatial immune architecture.

Raw Visium data were loaded using Scanpy~\citep{scanpy}. After reading the expression matrix and spatial metadata, we retained genes expressed in at least 50 cells with a minimum total count of 10. Counts were normalized to 1 million total transcripts per spot, and the top 2,000 HVGs were retained as input features for model training. The matrix was scaled and projected by PCA, retaining 200 components for downstream modeling.
\section{Results}\label{sec3}

% \subsection{Datasets}\label{subsec2}
% In this work, we employed spatial transcriptomic data generated with the 10× Genomics Visium platform. This technology captures transcripts within 55 µm barcoded spots, each covering multiple neighboring cells, and thus provides reliable molecular profiles at tissue scale. Two representative datasets were analyzed. The human dorsolateral prefrontal cortex (hDLPFC) ~\cite{dlpfc_1} dataset includes twelve tissue sections from three healthy donors, with each section comprising approximately 3,500–4,800 spots that were manually assigned to six cortical layers and white matter. The human breast cancer (hBrCa) ~\cite{SEDR} dataset consists of a single tumor section with 3,798 spots, reflecting both malignant epithelial regions and diverse immune infiltrates, which makes it suitable for investigating tumor architecture and microenvironmental interactions.

% \subsection{Clustering}\label{subsec2}
\subsection{Spatial domain identification}

\begin{table*}[ht]
\centering
\scriptsize
\renewcommand{\arraystretch}{0.9}
\setlength{\tabcolsep}{1pt} 
\caption{Comparison of clustering performance across methods on multiple datasets using ARI, AMI, and Completeness (COM) metrics. The best score
for each dataset is highlighted in \textbf{bold}, and the second-best
score is \underline{underlined}.}
% \begin{tabularx}{\textwidth}{@{}ll*{9}{>{\centering\arraybackslash}X}@{}}
\begin{tabularx}{\textwidth}{@{}lll*{9}{>{\centering\arraybackslash}X}@{}}

\toprule
Metric & Tissue  & Dataset& BayesSpace & conST & SEDR & Seurat & SpaceFlow & SpaGCN & STAGATE & stLearn & MultiST(ours) \\
\midrule
\multirow{13}{*}{ARI}
& Brain & 151507 & 0.435 & 0.340 & \underline{0.499} & 0.362 & 0.469 & 0.469 & \textbf{0.586} & 0.493 & 0.484 \\
& Brain &151508 & 0.421 & 0.309 & 0.453 & 0.441 & 0.313 & 0.354 & \underline{0.546} & 0.315 & \textbf{0.596} \\
& Brain &151509 & 0.343 & 0.305 & \underline{0.519} & 0.250 & 0.275 & 0.481 & \textbf{0.542} & 0.414 & 0.478 \\
& Brain &151510 & 0.429 & 0.259 & 0.450 & 0.413 & 0.243 & 0.445 & \underline{0.471} & 0.444 & \textbf{0.532} \\
& Brain &151669 & 0.437 & 0.300 & \textbf{0.519} & 0.357 & 0.309 & 0.099 & \underline{0.511} & 0.326 & 0.476 \\
& Brain &151670 & \underline{0.402} & 0.357 & 0.342 & 0.228 & 0.092 & 0.371 & \textbf{0.408} & 0.228 & 0.384 \\
& Brain &151671 & \textbf{0.711} & 0.475 & 0.572 & 0.441 & 0.273 & 0.503 & 0.589 & 0.389 & \underline{0.590} \\
& Brain &151672 & 0.562 & 0.483 & 0.569 & 0.373 & 0.404 & 0.563 & \underline{0.565} & 0.347 & \textbf{0.764} \\
& Brain &151673 & 0.547 & 0.517 & 0.511 & 0.424 & 0.391 & 0.461 & \underline{0.584} & 0.305 & \textbf{0.620} \\
& Brain &151674 & 0.280 & 0.468 & \textbf{0.610} & 0.379 & 0.340 & 0.323 & 0.381 & 0.386 & \underline{0.608} \\
& Brain &151675 & 0.532 & 0.394 & \textbf{0.625} & 0.303 & 0.314 & 0.300 & \underline{0.597} & 0.384 & 0.546 \\
& Brain &151676 & 0.371 & 0.478 & \underline{0.556} & 0.312 & 0.355 & 0.345 & 0.440 & 0.400 & \textbf{0.581} \\
& Breast &BRCA & 0.547 & 0.424 & 0.412 & 0.468 & 0.459 & \textbf{0.573} & 0.507 & 0.541 & \underline{0.552} \\
\midrule
\multirow{13}{*}{AMI}
& Brain &151507 & 0.604 & 0.483 & \underline{0.676} & 0.453 & 0.568 & 0.588 & \textbf{0.713} & 0.646 & 0.637 \\
& Brain &151508 & 0.568 & 0.450 & 0.635 & 0.460 & 0.484 & 0.459 & \underline{0.680} & 0.529 & \textbf{0.684} \\
& Brain &151509 & 0.547 & 0.448 & \textbf{0.673} & 0.378 & 0.473 & 0.607 & \underline{0.668} & 0.608 & 0.634 \\
& Brain & 151510 & 0.601 & 0.382 & \underline{0.633} & 0.488 & 0.429 & 0.554 & 0.602 & 0.598 & \textbf{0.664} \\
& Brain &151669 & 0.592 & 0.565 & 0.614 & 0.422 & 0.482 & 0.293 & \textbf{0.631} & 0.501 & \underline{0.626} \\
& Brain &151670 & \underline{0.550} & 0.513 & 0.518 & 0.362 & 0.349 & 0.486 & \textbf{0.573} & 0.410 & 0.549 \\
& Brain &151671 & 0.681 & 0.557 & \textbf{0.689} & 0.505 & 0.441 & 0.612 & \underline{0.688} & 0.549 & 0.679 \\
& Brain &151672 & 0.670 & 0.631 & \underline{0.685} & 0.418 & 0.522 & 0.652 & 0.683 & 0.491 & \textbf{0.747} \\
& Brain &151673 & 0.687 & 0.686 & 0.656 & 0.500 & 0.537 & 0.625 & \underline{0.719} & 0.497 & \textbf{0.731} \\
& Brain &151674 & 0.474 & 0.649 & \textbf{0.731} & 0.443 & 0.444 & 0.489 & 0.487 & 0.551 & 0.\underline{718} \\
& Brain &151675 & 0.696 & 0.545 & \textbf{0.731} & 0.412 & 0.451 & 0.458 & \underline{0.707} & 0.563 & 0.677 \\
& Brain &151676 & 0.567 & 0.623 & \underline{0.692} & 0.415 & 0.471 & 0.537 & 0.586 & 0.571 & \textbf{0.704} \\
& Breast &BRCA & \underline{0.671} & 0.590 & 0.660 & 0.627 & 0.675 & 0.670 & 0.654 & 0.645 & \textbf{0.672} \\
\midrule
\multirow{13}{*}{COM}
& Brain &151507 & 0.624 & 0.471 & 0.687 & 0.473 & 0.560 & 0.580 & \underline{0.698} & 0.677 & \textbf{0.703} \\
& Brain &151508 & 0.552 & 0.440 & 0.624 & 0.479 & 0.467 & 0.446 & \underline{0.660} & 0.560 & \textbf{0.799} \\
& Brain &151509 & 0.523 & 0.413 & \underline{0.663} & 0.366 & 0.435 & 0.601 & 0.636 & 0.612 & \textbf{0.679} \\
& Brain &151510 & 0.585 & 0.357 & \underline{0.601} & 0.494 & 0.398 & 0.536 & 0.581 & 0.584 & \textbf{0.649} \\
& Brain &151669 & 0.557 & 0.506 & 0.565 & 0.433 & 0.438 & 0.261 & \underline{0.586} & 0.478 & \textbf{0.597} \\
& Brain &151670 & \underline{0.490} & 0.452 & 0.458 & 0.329 & 0.307 & 0.424 & \textbf{0.505} & 0.379 & \underline{0.490} \\
& Brain &151671 & \textbf{0.682} & 0.542 & 0.653 & 0.483 & 0.413 & 0.578 & \underline{0.662} & 0.524 & 0.650 \\
& Brain &151672 & 0.649 & 0.628 & \underline{0.661} & 0.420 & 0.503 & 0.629 & 0.656 & 0.498 & \textbf{0.793} \\
& Brain &151673 & 0.674 & 0.676 & 0.651 & 0.522 & 0.522 & 0.609 & \underline{0.699} & 0.518 & \textbf{0.749} \\
& Brain &151674 & 0.467 & 0.639 & \textbf{0.728} & 0.461 & 0.440 & 0.480 & 0.475 & 0.562 & \underline{0.713} \\
& Brain &151675 & 0.712 & 0.541 & \textbf{0.746} & 0.427 & 0.443 & 0.452 & 0.703 & 0.583 & \underline{0.729} \\
& Brain &151676 & 0.567 & 0.610 & \underline{0.698} & 0.417 & 0.461 & 0.525 & 0.584 & 0.577 & \textbf{0.738} \\
& Breast &BRCA & 0.651 & 0.575 & 0.651 & 0.624 & 0.650 & \textbf{0.668} & 0.637 & 0.648 & \underline{0.663} \\
\bottomrule
\end{tabularx}
\vspace{-12pt}
\end{table*}

\begin{figure*}
    \centering
    \includegraphics[width=\linewidth]{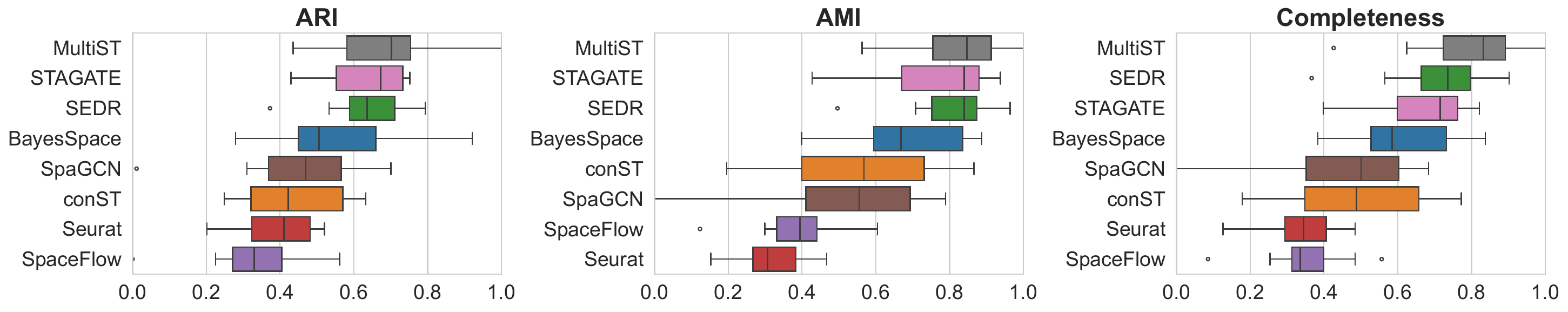}
    \vspace{-18pt}
    \caption{\textbf{MultiST consistently outperforms SOTA methods in clustering accuracy across 13 datasets.}
 % MultiST shows consistently higher accuracy and stability compared with other SOTA methods, including STAGATE, SEDR, BayesSpace, SpaGCN, conST, Seurat, and SpaceFlow.
 }
    \label{fig:clustering_metrics}
    \vspace{-15pt}
\end{figure*}

We first benchmarked clustering performance across 13 annotated datasets (12 DLPFC sections and one breast cancer section). As summarized in Fig.~\ref{fig:clustering_metrics}, MultiST consistently outperforms state-of-the-art (SOTA) methods in all three external evaluation metrics (ARI~\citep{ARI}, AMI~\citep{AMI}, Completeness~\citep{COM}), achieving both higher median scores and narrower interquartile ranges across datasets. To illustrate these results in detail, we next present two representative cases: DLPFC section 151673 and the breast cancer section.
% \begin{figure*}
%     \centering
%     \includegraphics[width=\linewidth]{image/clustering_metric_MultiST.pdf}
%     \caption{\textbf{MultiST consistently outperforms existing methods in clustering accuracy across 13 datasets.}
% Clustering performance was evaluated on 12 human dorsolateral prefrontal cortex (DLPFC) slices and one breast cancer (BRCA) sample. Boxplots summarize the distribution of  Adjusted Rand Index (ARI),  Adjusted Mutual Information (AMI), and  Completeness. MultiST shows consistently higher accuracy and stability compared with other SOTA methods, including STAGATE, SEDR, BayesSpace, SpaGCN, conST, Seurat, and SpaceFlow.}
%     \label{fig:clustering_metrics}
% \end{figure*}

% Clustering performance was evaluated on 12 human dorsolateral prefrontal cortex (DLPFC) slices and one breast cancer (BRCA) sample. Boxplots summarize the distribution of  Adjusted Rand Index (ARI),  Adjusted Mutual Information (AMI), and  Completeness.
% \begin{figure*}
%     \centering
%     \begin{minipage}[t]{0.6\linewidth}
%         \centering
%         \includegraphics[width=\linewidth]{image/cluster_pdf/151673_GT.pdf}
%         \textbf{(A)} 
%     \end{minipage}
%     \hfill
%     \begin{minipage}[t]{\linewidth}
%         \centering
%         \includegraphics[width=\linewidth]{image/cluster_pdf/combined_batch_151673.pdf}
%         \textbf{(B)} 
%     \end{minipage}
%     \caption{\textbf{Clustering results on the human DLPFC section 151673.} 
%     (A) Left: H\&E stained image. Right: Ground Truth annotation showing six cortical layers (L1–L6) and white matter (WM). 
%     (B) Comparison of clustering results from BayesSpace, conST, SpaceFlow, SEDR, Seurat, SpaGCN, STAGATE, and MultiST. 
%     For each method, spatial mapping (left), UMAP embedding (middle), and predicted clusters (right) are displayed. 
%     Adjusted Rand Index (ARI) values are provided to quantify consistency with manual annotations.
%     }
%     \label{fig:dlpfc_151673}
% \end{figure*}

\subsubsection{Clustering on human dorsolateral prefrontal cortex (DLPFC) }

% Among the 12 DLPFC sections, section 151673 is shown (Fig.S1) as a representative example due to its well-defined laminar structure (Layers 1--6 and WM), which makes it a standard benchmark for spatial clustering. 
% Results on the remaining 11 sections are provided in Supplementary Figures (Fig.S2--S12).

% DLPFC section 151673 exhibits the canonical laminar organization of the cortex, consisting of Layers~1--6 (L1--L6) and the underlying white matter (WM). Each layer has distinct anatomical and functional properties: L1 is sparsely populated, dominated by dendrites and axons; L2/3 contain numerous pyramidal neurons supporting cortico-cortical communication; L4 is the principal recipient of thalamic inputs; L5 hosts large pyramidal neurons projecting to subcortical targets; L6 participates in corticothalamic feedback; and WM comprises mainly myelinated axons. An ideal spatial clustering method should therefore recover the global arrangement of L1-L6 and WM while maintaining sharp boundaries at critical transitions (e.g., L6--WM) and sufficient resolution for fine-grained layers such as L4.

Among the 12 DLPFC sections, section 151673 (Fig.~S1) was selected as a representative benchmark due to its well-defined laminar organization, consisting of cortical layers L1--L6 and the underlying white matter (WM). The DLPFC exhibits a complex laminar architecture with subtle transcriptional differences between adjacent layers, making accurate spatial clustering particularly challenging. An effective method should recover the global laminar layout while preserving sharp inter-layer boundaries (e.g., L6--WM) and sufficient resolution for fine-grained layers such as L4. Results for the remaining 11 sections are provided in supplementary material Figs.~S2--S12.

As shown in Fig.~S1, existing methods exhibit several limitations. Some approaches (e.g., BayesSpace and SpaGCN) capture the overall laminar structure but produce jagged or ambiguous boundaries, particularly at the L6--WM interface. Methods lacking strong spatial constraints (e.g., Seurat) yield blurred boundaries and merge transcriptionally similar layers such as L2/3 and L4. Other methods (e.g., SEDR and SpaceFlow) tend to over-smooth spatial patterns, resulting in loss of fine-grained layer resolution. Even the strongest baseline, STAGATE, successfully restores the global laminar organization but still shows partial mixing around L4. Notably, delineating Layer 4 remains a common challenge across methods, reflecting its transcriptional similarity to neighboring layers and the spot-level resolution of Visium data.

In contrast, MultiST produces results most consistent with manual annotation. On the spatial map, it captures the L6--WM boundary sharply and preserves laminar continuity across the section. In the UMAP embedding, MultiST forms compact and well-separated layer-specific clusters, reducing mixing compared to other methods. Although some local ambiguity remains around L4, MultiST overall provides clearer layer boundaries and more consistent laminar patterns.

Quantitatively MultiST achieves the highest scores on all three metrics (ARI=0.620, AMI=0.731, Completeness=0.749). Compared with the strongest baseline STAGATE, MultiST improves ARI by 6.2\% and Completeness by 7.2\%. These gains confirm that MultiST reduces cross-layer misclassification and better preserves within-layer consistency.

In summary, DLPFC section 151673 demonstrates that existing approaches suffer from limitations in boundary sharpness, laminar continuity, or fine-grained resolution. MultiST, in contrast, provides clearer layer separation and higher biological consistency. Results on the remaining 11 DLPFC sections, provided in the Supplementary Figures, show consistent trends and further confirm the robustness of MultiST across samples.

\subsubsection{Clustering on human breast cancer (BRCA)}

\begin{figure*}[p]
\centering
\captionsetup{font=small, width=0.92\textwidth, skip=6pt, justification=raggedright, singlelinecheck=false}
\includegraphics[width=\linewidth]{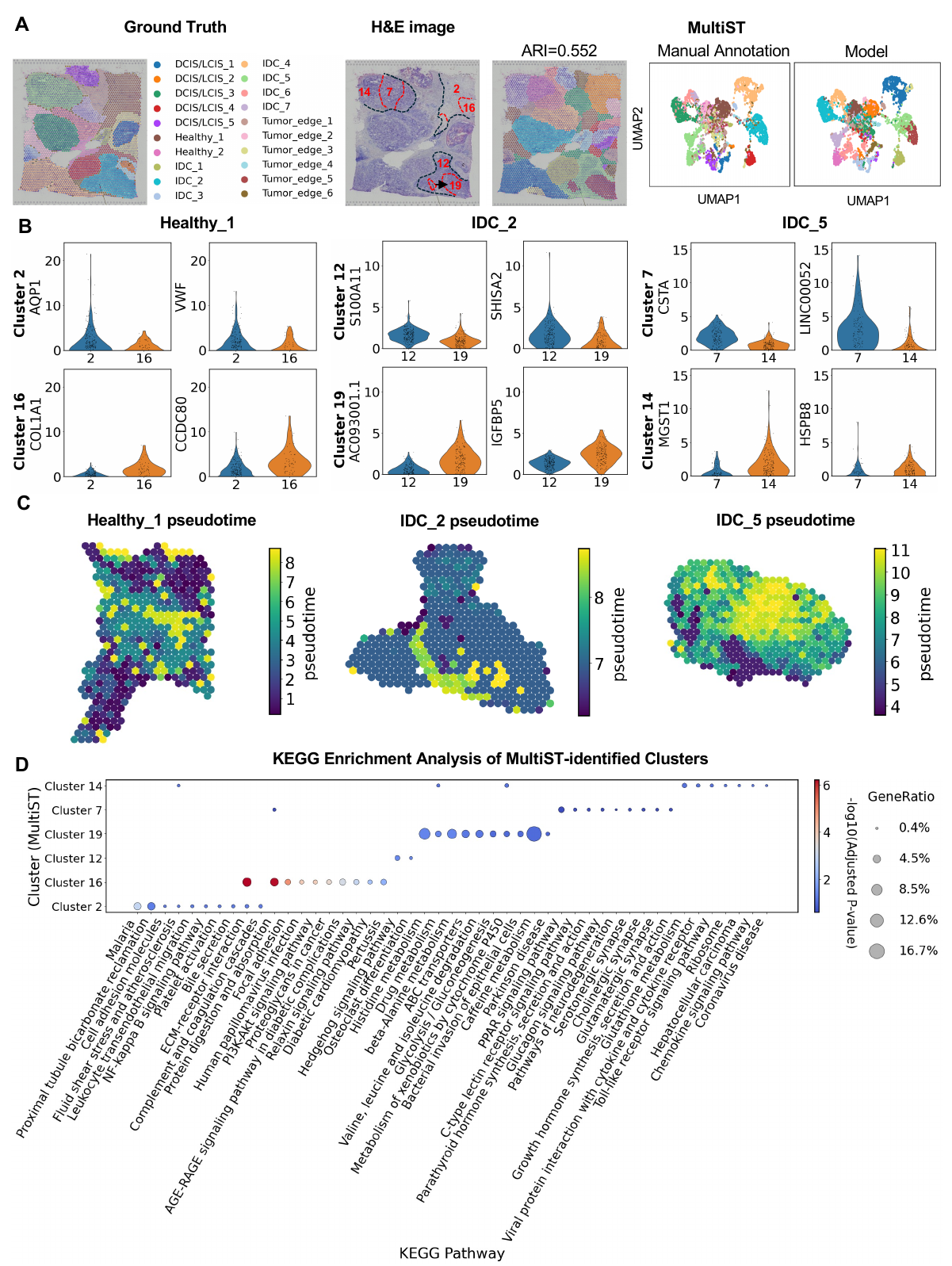}
\vspace{-18pt}
\caption{
\textbf{Biological validation of MultiST-based spatial clustering in breast cancer tissue sections.}
(A) Ground truth, H\&E histology, and MultiST-predicted spatial domains.
In the H\&E panel, black outlines indicate ground-truth tissue regions, while red numbers mark the dominant MultiST clusters. 
% In the H\&E panel, black outlines indicate histologically validated regions (Healthy\_1, IDC\_2, IDC\_5), while red numbers denote the two dominant MultiST clusters per region.
(B) Representative marker genes for selected clusters. 
(C) Spatial pseudotime maps inferred by Monocle3. %Spatial pseudotime maps of Healthy\_1, IDC\_2, and IDC\_5 reconstructed using Monocle3.
(D) KEGG enrichment analysis of MultiST clusters.
% , with dot size indicating \textit{GeneRatio} and color showing adjusted $-\log_{10}(p)$ values.
% \textbf{Biological validation of MultiST-based spatial clustering in breast cancer tissue sections.}
%     (A) Ground truth, H\&E histology, and MultiST-predicted spatial domains (ARI = 0.552).
% In the H\&E panel, black outlines show validated tissue regions (Healthy\_1, IDC\_2, IDC\_5), and red numbers label the two most abundant MultiST clusters per region.
%     (B) Violin plots showing representative marker genes for selected MultiST-identified clusters.
%     (C) Spatial pseudotime maps of Healthy\_1, IDC\_2, and IDC\_5 regions generated with Monocle3.
%     (D) KEGG pathway enrichment analysis of MultiST-identified clusters, with dot size indicating GeneRatio and color denoting adjusted $-\log_{10}(p)$ values.
}
\label{fig:BRCA}
\end{figure*}

In contrast to the laminar organization of DLPFC, breast cancer tissue exhibits pronounced spatial heterogeneity, comprising ductal carcinoma in situ (DCIS/LCIS), invasive ductal carcinoma (IDC), tumor edge regions, and residual healthy tissue. The dataset contains 20 annotated spatial domains, making accurate clustering particularly challenging.
An ideal clustering should preserve the integrity of large pathological domains, distinguish IDC from tumor edge compartments, and retain isolated DCIS/LCIS foci as independent clusters. In the low-dimensional embedding, these features are typically reflected by compact and well-separated clusters rather than fragmented or overlapping distributions.  

Among the baseline methods( Fig.~S13), conST, Seurat, and STAGATE generate relatively scattered partitions, with blurred spatial boundaries and insufficient separation in the UMAP embedding. SpaceFlow produces excessive fragmentation, resulting in numerous small clusters that fail to accurately properly separate IDC, tumor edge, and DCIS/LCIS. SEDR displays cross-domain merging, incorrectly grouping parts of tumor edge, IDC, and even healthy tissue into a single cluster, markedly deviating from pathological annotations. BayesSpace achieves a moderate ARI (0.547) and preserves some domain structures. However, it tends to over-segment regions into many small clusters, with irregular boundaries and boundary contamination. SpaGCN achieves comparatively better performance by recovering the major domains,  but its cluster boundaries remain irregular and subject to contamination by heterogeneous points.  

By comparison, MultiST provides more balanced and biologically interpretable results (Fig.~\ref{fig:BRCA}A). In the spatial domain, DCIS/LCIS regions are preserved as coherent units, and tumor edge compartments are more consistently delineated. In the UMAP embedding, IDC and tumor edge clusters appear compact and clearly separated, avoiding the fragmentation observed in Seurat and SEDR and the over-smoothing seen in SpaceFlow. Quantitatively, MultiST achieves ARI=0.552, AMI=0.672, and Completeness=0.663, ranking among the top across all three metrics. Although SpaGCN marginally outperforms in ARI, MultiST obtains the best AMI, indicating reduced fragmentation and greater overall consistency. Importantly, in cancer tissue where transcriptional differences between IDC and tumor edge are often subtle, the incorporation of histology-derived features provides complementary morphological cues. This likely explains why both MultiST and SpaGCN show advantages in boundary delineation, with MultiST yielding more stable and interpretable clustering overall.  

Taken together, these results demonstrate that MultiST preserves large-scale integrity, maintains clearer IDC–edge separation, and sustains the independence of DCIS/LCIS regions, while consistently performing well across external evaluation metrics. These clustering outcomes establish a reliable basis for subsequent biological validation. In the following sections, we integrate marker gene expression, KEGG pathway enrichment, pseudotime inference, and CCI analysis to further assess the biological relevance of the identified domains.

\subsubsection{Biological validation in breast cancer}
To further validate MultiST clustering in breast cancer, we focused on three
representative regions from the GT annotation: Healthy\_1, IDC\_2, and IDC\_5.
These regions were consistently separated into distinct clusters by MultiST,
whereas baseline methods often showed merging or blurred boundaries, making
them informative for validation.
Differential gene expression (DGE) analysis was conducted using Scanpy’s
\texttt{rank\_genes\_groups} function with the Wilcoxon rank-sum test in a
one-versus-rest manner. Genes with log$_2$ fold change (logFC) greater than
1.5 and Benjamini--Hochberg adjusted $p$-values below 0.05 were considered
significant. From the significant gene sets, top-ranked markers are selected
for visualization, and the results are examined by KEGG pathway enrichment
and pseudotime inference with Monocle3~\cite{Monocle3}
(Fig.~\ref{fig:BRCA}B--\ref{fig:BRCA}D).

Within Healthy\_1, MultiST resolves two molecularly distinct populations despite their shared annotation as normal tissue. One population is enriched for extracellular matrix--related genes such as \textit{COL1A1} and \textit{CCDC80}, whereas the other is characterized by vascular-associated genes including \textit{AQP1} and \textit{VWF}. Pseudotime analysis reveals a continuous transcriptional transition between these populations, suggesting microenvironmental variation within the GT-defined normal tissue region.

% The IDC\_2 region displays clear functional stratification. Cluster~12 exhibits elevated expression of \textit{S100A11} and \textit{SHISA2} and enrichment in Hedgehog signaling, consistent with a proliferative and invasive--associated transcriptional program. In contrast, Cluster~19 expresses \textit{AC093001.1} and \textit{IGFBP5} and is enriched for pathways related to viral infection, ABC transporters, and drug metabolism, indicating a stress- and metabolism-associated transcriptional state. These clusters are positioned along a continuous pseudotime trajectory, reflecting related but distinct transcriptional states within the same GT region.
IDC\_2 displays pronounced functional heterogeneity. Cluster 12 expresses \textit{S100A11} and \textit{SHISA2}, enriched in Hedgehog signaling, consistent with a proliferative and invasive IDC state. By contrast, Cluster 19 expresses \textit{AC093001.1} and \textit{IGFBP5}, enriched in virus infection, ABC transporters, and drug metabolism, reflecting a stress-adapted and metabolically reprogrammed IDC population. Pseudotime inference shows a continuous trajectory between these subgroups, indicating that IDC cells within the same GT region may occupy distinct functional states or progression stages. This proliferative–stress dichotomy illustrates MultiST’s ability to resolve intra-tumor heterogeneity beyond GT annotation.

Similarly, IDC\_5 is further divided into two divergent populations. 
Cluster 7 is characterized by expression of \textit{CSTA} and \textit{LINC00052} and shows enrichment in metabolic and signaling pathways such as PPAR signaling and glucagon signaling, suggesting a metabolically active IDC subtype.
Cluster 14, in contrast, is marked by the expression of \textit{MGST1} and \textit{HSPB8} and exhibits significant enrichment in chemokine signaling, Toll-like receptor signaling, and viral protein–cytokine interaction pathways, indicative of an immune–inflammatory subtype.
Pseudotime analysis reveals that these groups are distributed along a dynamic trajectory rather than completely isolated, suggesting the coexistence of proliferative and immune–inflammatory states within the same IDC region. This subdivision underscores MultiST’s ability to capture fine-grained heterogeneity in breast cancer and provides a biological rationale for downstream CCI analysis.

Overall, MultiST not only recapitulates the major spatial domains of the breast cancer section but also reveals functional heterogeneity within GT-defined regions. Both the microenvironmental differences in Healthy\_1 and the proliferative--stress--immune subtypes within IDC are supported by DEGs, KEGG enrichment, and pseudotime analysis. These findings provide strong biological support for the clustering performance of MultiST and set the stage for investigating tumor microenvironmental interactions in subsequent analyses.

\subsection{Pseudotime Trajectory Inference}\label{subsec2}
We next assessed the ability of different methods to reconstruct developmental or pathological trajectories from spatial transcriptomics data. Pseudotime inference provides insights into dynamic cellular processes by ordering spatial spots along continuous axes, thereby complementing discrete clustering results. Among the benchmarked datasets, we presented representative analyses on the DLPFC section 151673, which captures the laminar organization of the human cortex, and the breast cancer section, which reflects spatially heterogeneous tumor progression. Results  are provided in Supplementary Figures (Fig.~S14–S26).

% \begin{figure*}
%     \centering
%     \includegraphics[width=0.95\linewidth]{image/trajectory_pdf/all_models_151673_comparison_finalHD.pdf}
%     \caption{\textbf{Pseudotime trajectory inference on DLPFC section 151673.}
% Pseudotime was inferred using Monocle3 with embeddings from eight representative methods (BayesSpace, conST, SpaceFlow, SEDR, Seurat, SpaGCN, STAGATE, and MultiST). The white matter (WM) region was designated as the root, leading to pseudotime values increasing from WM toward superficial cortical layers. For each method, the left panel shows trajectories in UMAP space, while the right panel maps pseudotime values back to spatial coordinates.}
%     \label{fig:dlpfc_pseudotime}
% \end{figure*}

\subsubsection{DLPFC: Developmental Spatial Layer Mapping}
In the DLPFC section 151673, we compared the performance of different methods in both low-dimensional UMAP trajectories and spatial mappings (Fig.~S14).
Cortical development follows an “inside–out” pattern, with superficial layers (L1--L2/3) progressing through L4--L6 to WM~\citep{layers1,layers2,layers3}.
% Biologically, cortical development follows an “inside–out” pattern, with superficial layers (L1--L2/3) gradually progressing through L4 and L5 before reaching L6 and ultimately W~\citep{layers1,layers2,layers3}.
In Monocle3, WM was set as the root, so pseudotime values increase outward, consistent with the expected progression from superficial to deep layers.
% In Monocle3, we set WM as the root, such that pseudotime values numerically increase outward from WM, while the biological expectation remains a progression from superficial to deep layers. 

Among baseline methods, Seurat, BayesSpace, SpaceFlow, and SEDR produce circular trajectories in UMAP space, preventing a monotonic ordering. In these cases, WM is often incorrectly aligned to L1, L2, or L4 instead of L6, leading to layer inversion inconsistent with cortical lamination. 
% Their spatial maps also fail to recover a clear superficial–deep gradient. For instance, Seurat and BayesSpace fragment WM into discontinuous parts, while cortical layers do not exhibit a coherent sequential relationship. 
Their spatial maps fail to capture a clear superficial-to-deep gradient; Seurat and BayesSpace fragmented WM, and cortical layers lacked sequential coherence.
conST generates smoother spatial embeddings but still displays discontinuities and irregular jumps in pseudotime gradients. SpaGCN partially restores the superficial-to-deep continuity, yet transitions at critical boundaries (e.g., L5--L6 and L6--WM) remain blurred, with locally inconsistent pseudotime values. STAGATE yields relatively more reasonable trajectories, recovering an overall gradient from superficial layers to WM, although the transitions near WM remain unstable.  

In contrast, MultiST produces trajectories most consistent with manual annotations. 
Its UMAP embeddings reveal a single continuous backbone, and spatial mappings showed a smooth progression from superficial layers through L6 to WM, with sharper L6--WM transitions.
% Its UMAP embeddings reveal a single continuous backbone, avoiding circular structures, while spatial mappings exhibit a smooth progression from superficial layers through L6 to WM, with sharper transitions at the L6--WM interface. 
These results demonstrate that by integrating histology with gene expression, MultiST provides more stable and biologically interpretable embeddings, enabling more accurate reconstruction of cortical laminar trajectories.

\subsubsection{BRCA: Disease Progression Mapping}
% \begin{figure*}
%     \centering
%     \includegraphics[width=0.9\linewidth]{image/trajectory_pdf/all_models_V1_Human_Breast_Cancer_Block_A_Section_1_comparison_finalHD.png}
%     \caption{\textbf{Pseudotime trajectory inference on breast cancer tissue.}
% Pseudotime trajectories were inferred with Monocle3 using low-dimensional embeddings from eight methods (BayesSpace, conST, SpaceFlow, SEDR, Seurat, SpaGCN, STAGATE, and MultiST). The Healthy\_1 region was designated as the root to approximate the pathological progression from healthy tissue (H) through ductal carcinoma in situ (DCIS/LCIS, D) and tumor edge (T), ultimately reaching invasive ductal carcinoma (IDC, I). For each method, the left panel shows pseudotime trajectories in UMAP space, and the right panel presents pseudotime values mapped back to spatial coordinates.}
%     \label{fig:placeholder}
% \end{figure*}
% In breast cancer sections, the goal of pseudotime analysis differs from that in the DLPFC. Whereas the DLPFC emphasizes laminar developmental ordering, breast cancer lacks fixed anatomical layers, and the focus is delineating gradual transitions between pathological states. An ideal pseudotime trajectory should reflect progression from healthy tissue (Healthy, H), through ductal carcinoma in situ (DCIS/LCIS, D) and tumor edge (T), ultimately reaching invasive ductal carcinoma (IDC, I), corresponding in space to a smooth gradient from the periphery toward the tumor core. Using Monocle3, we set  Healthy\_1 (H1) as the root to approximate this direction.

In breast cancer section, pseudotime analysis serves a different goal than in
the DLPFC. Unlike cortical layers, breast cancer lacks fixed
anatomical structures, and the focus is delineating
transitions between pathological states. An ideal trajectory
should reflect progression from healthy tissue (Healthy, H),
through DCIS/LCIS (D) and tumor edge (T), to invasive ductal
carcinoma (IDC, I), corresponding spatially to a gradient from
tissue periphery toward the tumor core. Using Monocle3,
Healthy 1 (H1) was set as the root to approximate this
direction.

MultiST pseudotime mapping reveals fine-grained spatial patterns and a continuous gradient across the tissue (Fig.~S26). Pseudotime smoothly increases from peripheral healthy regions to invasive carcinoma, capturing transitional tumor edge regions (T1--T6) that reflect progressive malignant states. Adjacent regions display consistent pseudotime values, preserving spatial continuity and accurately representing the gradual distribution of malignant potential. Compared with baseline methods, MultiST demonstrates greater biological plausibility: SEDR exhibits extreme distributions, Seurat shows abrupt jumps, BayesSpace is ambiguous at DCIS--IDC boundaries, SpaceFlow displays patchy patterns, and conST generates rigid boundaries. SpaGCN and STAGATE partially recover the global gradient but suffer from over-smoothing or complex multilayer structures. Overall, MultiST provides a more interpretable global pseudotime gradient, highlighting major spatial patterns of malignant progression and supporting downstream analyses.

Pseudotime trajectories projected in UMAP space from MultiST show a network-like, dispersed configuration rather than a linear backbone. This reflects both transcriptional heterogeneity and UMAP's limited ability to order pathological states, so UMAP is better interpreted as visualizing transcriptional similarity than temporal progression. In the MultiST projection: (1) healthy tissues H1 and H2 cluster closely but lack a distinct starting point; (2) IDC regions I1--I7 are widely scattered, highlighting high heterogeneity; and (3) the central network shows mixed connections between H1/H2 and tumor edge regions T1/T3, indicating overlapping gene expression. Similar patterns are seen in other methods (SpaGCN, BayesSpace, STAGATE). Ground Truth supports the biological plausibility: tumor edges T1/T3 surround H1--H2, and spatial proximity generates microenvironmental similarities reflected in UMAP. Overall, MultiST preserves transcriptional heterogeneity and spatial continuity, facilitating interpretation of tumor organization and microenvironmental features.

\subsection{Cell–Cell Interaction Analysis}\label{subsec2}

% \begin{figure*}
%     \centering
%     \begin{minipage}[t]{\linewidth}
%         \centering
%         \includegraphics[width=\linewidth]{image/CCI.pdf}
%         \textbf{(A)} 
%     \end{minipage}
%     \hfill
%     \begin{minipage}[t]{\linewidth}
%         \centering
%         \includegraphics[width=\linewidth]{image/spateo.pdf}
%         \textbf{(B)} 
%     \end{minipage}
%     % \begin{minipage}[t]{\linewidth}
%     %     \centering
%     %     \includegraphics[width=\linewidth]{image/spateo.pdf}
%     %     \textbf{(C)} 
%     % \end{minipage}
%     \caption{\textbf{Cell–cell communication analysis in breast cancer tissue. }
% (A) Heatmap of overall communication strength between the four major regions (left), and barplot showing the top ligand--receptor interactions with pathway annotation (right). 
% (B) Bubble plot of the top 20 ligand–target interactions across region pairs, where rows represent source–target region pairs and columns denote ligand–target gene pairs. The color intensity indicates the number of significant interactions.
% }
  
    % \label{fig:CCI}
% \end{figure*}

\subsubsection{Ligand--Receptor interaction network analysis}
% After completing pseudotime inference, we further performed CCI analysis on the BRCA based on the partitions identified by our model. This choice is biologically meaningful, as tumors are now recognized not merely as aggregates of malignant cells but as complex ecosystems composed of cancer cells and their surrounding microenvironment~\citep{cancerSystem1,cancerSystem2}. The expansion and survival of cancer cells rely not only on intrinsic genetic alterations but also on their ability to recruit and reprogram immune cells, fibroblasts, and endothelial cells through secreted signaling molecules~\citep{microenvironment1,microenvironment2}, thereby establishing supportive communication networks that ensure nutrient supply, immune evasion, and invasive capacity~\citep{crosstalk1,crosstalk2}.

% To enhance interpretability and statistical power, the 20 original clusters were consolidated into four major categories: DCIS/LCIS, IDC , Healthy , and Tumor  edge . This integration strategy simplifies the biological hierarchy and increases the sample size of each category, making interaction patterns more robust. Subsequently, CellChat~\citep{cellchat} was applied to systematically infer ligand--receptor interactions. By relying on model-derived clusters rather than predefined annotations, this approach ensures that the CCI analysis directly reflects the spatial structures identified by our method, thereby validating its biological relevance.
After pseudotime inference, we performed cell–cell interaction (CCI) analysis on the breast cancer section using the partitions identified by MultiST. This approach is biologically meaningful, as tumors are increasingly recognized not merely as aggregates of malignant cells but as complex ecosystems comprising cancer cells and their surrounding microenvironment~\citep{cancerSystem1,cancerSystem2}. The expansion and survival of cancer cells depend not only on intrinsic genetic alterations but also on their capacity to recruit and reprogram immune cells, fibroblasts, and endothelial cells via secreted signaling molecules~\citep{microenvironment1,microenvironment2}, thereby establishing supportive communication networks that facilitate nutrient supply, immune evasion, and invasive potential~\citep{crosstalk1,crosstalk2}. To enhance interpretability and statistical power, the 20 original clusters were consolidated into four major categories—DCIS/LCIS, IDC, Healthy, and Tumor edge—simplifying the biological hierarchy and increasing sample size for more robust interaction patterns. Subsequently, CellChat~\citep{cellchat} was applied to systematically infer ligand--receptor interactions, and by relying on model-derived clusters rather than predefined annotations, this analysis directly reflects the spatial structures identified by MultiST, thereby validating their biological relevance. 

\begin{figure*}
    \centering
    \includegraphics[width=\linewidth]{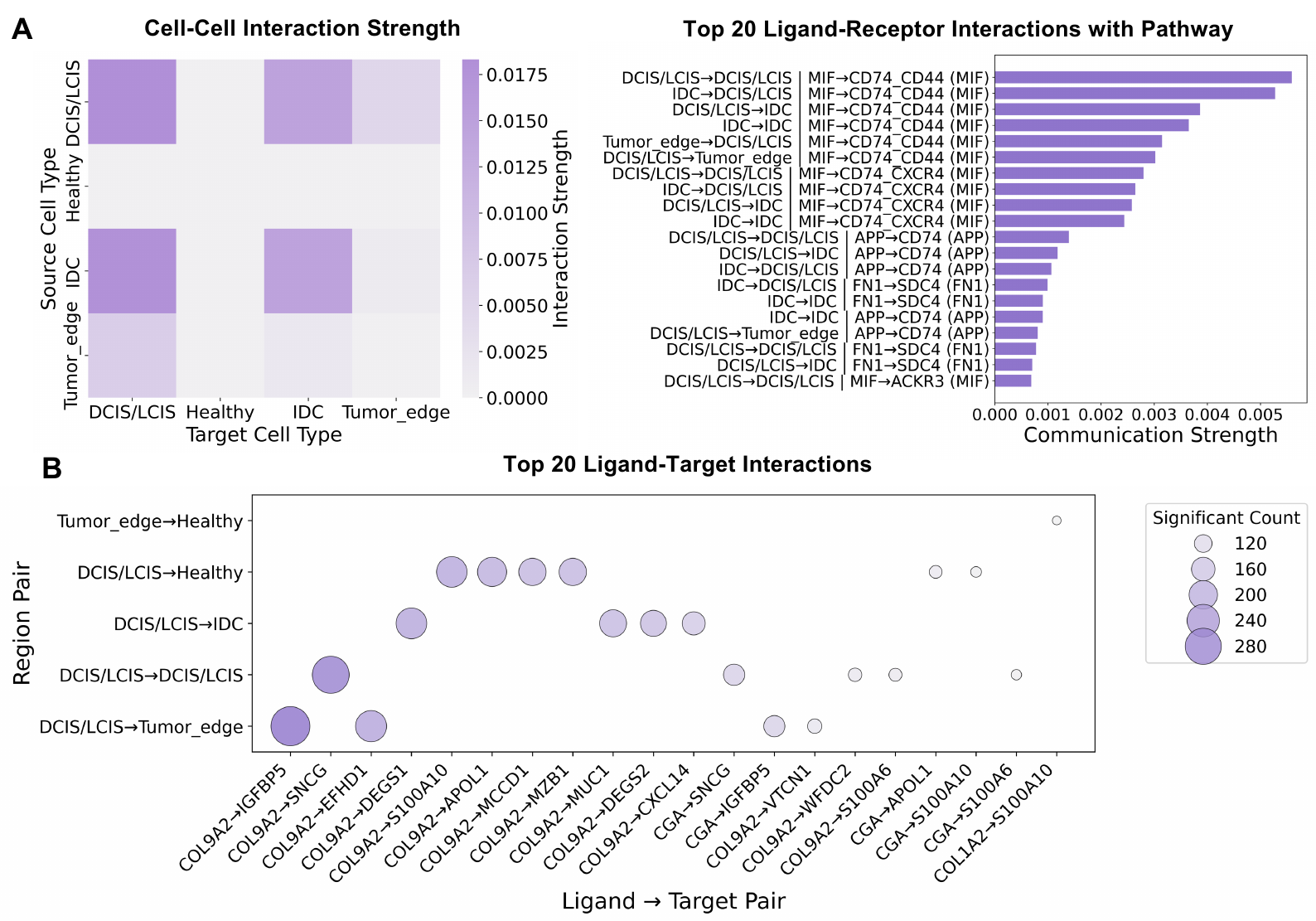}
    \vspace{-18pt}
    \caption{\textbf{Cell–cell communication analysis in breast cancer tissue. }(A) Heatmap of overall communication strength between major regions (left) and barplot of top ligand--receptor interactions with pathway annotation (right). (B) Bubble plot showing significant ligand–target interactions across region pairs, with color intensity reflecting interaction counts.}
% (A) Heatmap of overall communication strength between the four major regions (left), and barplot showing the top ligand--receptor interactions with pathway annotation (right). 
% (B) Bubble plot of the top 20 ligand–target interactions across region pairs, where rows represent source–target region pairs and columns denote ligand–target gene pairs. The color intensity indicates the number of significant interactions.}
    \label{fig:CCI}
    \vspace{-15pt}
\end{figure*}

Fig.~\ref{fig:CCI}A (left) illustrates the overall communication strengths among the four categories. The most feature is the strong self-interaction within DCIS/LCIS, far exceeding all other types, suggesting dense autocrine and paracrine signaling networks that may reinforce proliferative and survival capacities through positive feedback. This is followed by DCIS/LCIS--IDC cross-interactions and IDC self-interactions, indicating highly active communication between early and invasive stages, which may reflect reciprocal support during malignant progression. In contrast, tumor edge exhibits relatively weak interactions with all categories, possibly due to its lower cell density or unstable transitional status, suggesting a more passive role as a signal receiver rather than a dominant sender.
% Fig.~\ref{fig:CCI}A(left) illustrates the overall communication strengths among the four categories. The most striking feature is the strong self-interaction within DCIS/LCIS, far exceeding all other types, suggesting the presence of dense autocrine and paracrine signaling networks that may reinforce proliferative and survival capacities through positive feedback. This is followed by DCIS/LCIS → IDC cross-interactions and IDC self-interactions, indicating highly active communication between early and invasive stages, which may reflect reciprocal support during malignant progression. In contrast, tumor edge exhibits relatively weak interactions with all categories, possibly due to its lower cell density or unstable transitional status, suggesting a more passive role as a signal receiver rather than a dominant sender.

Fig.~\ref{fig:CCI}A (right) presents the top-ranked ligand--receptor pairs. The \textit{MIF--CD74/CD44} axis dominates the interaction landscape, with the top five interactions originating from this pathway. By binding to \textit{CD74} and co-receptor \textit{CD44}, \textit{MIF} activates \textit{ERK1/2} and \textit{AKT} signaling cascades, promoting cancer cell proliferation and survival; in parallel, it induces epithelial--mesenchymal transition to enhance invasiveness, suppresses T-cell function to facilitate immune escape, and promotes angiogenesis~\citep{CD44/CD74_1,CD44/CD74,MIF}. These results suggest that the \textit{MIF--CD74/CD44} axis is a prominent communication that can help maintain the malignant phenotype of breast cancer, enabling reciprocal support between different tumor regions. Additional pathways, such as \textit{MIF--CD74/CXCR4}, \textit{APP--CD74}, and \textit{FN1--SDC4}, are enriched, highlighting roles in chemotaxis, stress responses, and extracellular matrix remodeling.

% \citep{CD44/CD74_1,CD44/CD74,MIF}

\subsubsection{Ligand--Target interaction network analysis}
% To further elucidate the molecular mechanisms underlying cell--cell communication across regions with distinct malignant states, we integrate ligand--receptor interactions identified by CellChat with differentially expressed genes inferred from the Spateo model~\citep{spateo}. Based on breast cancer expression profiles and curated ligand--receptor databases, we select representative ligands and targets expressed across four tissue categories (DCIS/LCIS, IDC, Healthy, and Tumor edge). We then prioritize the top 20 ligand--target pairs according to their interaction significance (Significant Count), which represents the number of cells within each cell type for which the ligand–target interaction z-score exceeds 1.96, indicating statistically significant association (Fig.~\ref{fig:CCI}B).
% To further elucidate molecular mechanisms of cell–cell communication across regions with distinct malignant states, we integrated ligand–receptor interactions from CellChat with differentially expressed genes inferred by the Spateo model~\citep{spateo}. Based on breast cancer expression profiles and curated ligand–receptor databases, we selected representative ligands and targets across four tissue categories (DCIS/LCIS, IDC, Healthy, and Tumor edge). The top 20 ligand–target pairs were prioritized by Significant Count, defined as the number of cells within each cell type where the ligand–target interaction z-score exceeds 1.96 (Fig.~\ref{fig:CCI}B).
To further elucidate mechanisms of cell--cell communication across regions with distinct malignant states, we integrated ligand--receptor interactions from CellChat with differentially expressed genes inferred by \textit{Spateo}~\citep{spateo}. Based on breast cancer expression profiles and curated ligand--receptor databases, we selected representative ligands and targets across four tissue categories (DCIS/LCIS, IDC, Healthy, and Tumor edge). The top 20 ligand--target pairs were prioritized by Significant Count, defined as number of cells within each cell type where the ligand--target interaction z-score~\citep{zscore} exceeds 1.96 (Fig.~\ref{fig:CCI}B).

% The resulting heatmap highlights five dominant communication patterns: intra-DCIS/LCIS signaling, DCIS/LCIS→Healthy, DCIS/LCIS→IDC, DCIS/LCIS→Tumor\_edge, and Tumor\_edge→Healthy cross-regional signaling.

% Network-level interrogation of ligand--target pairs uncovers molecular mechanisms that underpin tumor progression. COL9A2, an extracellular matrix (ECM) component, emerges as a dominant hub ligand. Notably, the \textit{COL9A2}→\textit{SNCG} interaction appears with the highest frequency among the top-ranked pairs. \textit{SNCG} ($\gamma$-synuclein), initially identified as a breast cancer–specific gene, is consistently upregulated in invasive and metastatic disease and has been recognized as a diagnostic and prognostic marker~\citep{SNCG}. Its recurrent interaction with \textit{COL9A2} suggests that DCIS/LCIS regions may not only remodel the ECM but also directly activate pro-metastatic transcriptional programs, thereby facilitating the transition from in situ to invasive carcinoma. This highlights a critical molecular event in which ECM-derived signals drive malignant phenotypic conversion.

Network-level interrogation of ligand--target pairs uncovers molecular mechanisms underpinning tumor progression. \textit{COL9A2}, an extracellular matrix (ECM) component, emerges as a dominant hub ligand. The \textit{COL9A2} $\rightarrow$ \textit{SNCG} interaction appears with the highest frequency among top-ranked pairs. \textit{SNCG} ($\gamma$-synuclein), initially identified as a breast cancer-specific gene, is consistently upregulated in invasive and metastatic disease and recognized as a diagnostic and prognostic marker~\citep{SNCG}. The recurrent coupling of COL9A2 with SNCG suggests that ECM-remodeling programs in DCIS/LCIS may be accompanied by an SNCG-high pro-invasive state, potentially facilitating progression toward invasive carcinoma. %Its recurrent coupling of \textit{COL9A2} suggests that DCIS/LCIS regions not only remodel the ECM but also directly activate pro-metastatic transcriptional programs, facilitating the transition from in situ to invasive carcinoma. This highlights a critical molecular event where ECM-derived signals drive malignant phenotypic conversion.
% Beyond \textit{SNCG}, \textit{COL9A2} engages a spectrum of functional targets that expand the pro-tumor signaling network. Interactions with \textit{S100A10}, \textit{S100A6}, and \textit{EFHD1} implicate calcium-dependent pathways and cytoskeletal remodeling in enhanced cellular motility. The \textit{COL9A2}→\textit{DEGS1/2} axis connects ECM remodeling with sphingolipid metabolism, potentially reshaping the balance between cell survival and apoptosis. Similarly, the \textit{COL9A2}→\textit{APOL1} pathway may couple ECM signaling to lipid transport and inflammatory responses, underscoring the integration of metabolic and immune cues into tumor--microenvironment communication.

Beyond \textit{SNCG}, \textit{COL9A2} associated ligan-target links include \textit{S100A10}, \textit{S100A6}, and \textit{EFHD1}, suggesting enrichment of calcium-dependent cytoskeletal remodeling programs that may support motility. The inferred \textit{COL9A2}→\textit{DEGS1/2} axis suggests coupling between ECM remodeling and sphingolipid/ceramide metabolism, potentially influencing the balance between survival and apoptosis. Similarly, the \textit{COL9A2}→\textit{APOL1} association may connect ECM-linked states with lipid-handling and inflammatory signaling, highlighting metabolic and immune cues in tumor–microenvironment communication.

Two additional axes show potential therapeutic relevance. First, within IDC, the AGRN→DAG1 ligand–receptor interaction implicates dystroglycan-mediated ECM adhesion/signaling, this axis may reflect an adhesion/mechanotransduction state permissive for invasion~\citep{AGRN,DAG1}. S100A11, which has been associated with invasive behavior and immune-checkpoint–linked programs, is concurrently elevated, suggesting coupling between AGRN–DAG1 signaling and an S100A11-high invasive/immune-evasive state~\citep{S100A11,S100}.%the \textit{AGRN}→\textit{DAG1}→\textit{S100A11} pathway: agrin engages dystroglycan (\textit{DAG1}) to reinforce adhesion and migration~\citep{AGRN,DAG1}, while \textit{S100A11} has been linked to invasion and immune suppression~\citep{S100A11,S100}. 
The strong activity of this axis in IDC suggests that it sustains both the invasive phenotype and an immunosuppressive niche, and its disruption may simultaneously attenuate invasion and restore anti-tumor immunity~\citep{ECM_1,ECM_2}. Second, the inferred \textit{COL9A2}→\textit{SDC1}→\textit{ALDH3B2} axis from DCIS/LCIS to the tumor edge: syndecan-1 (\textit{SDC1}) mediates ECM--receptor signaling and adhesion~\citep{SDC1}, 
whereas \textit{ALDH3B2}, a member of the aldehyde dehydrogenase family implicated in cancer cell plasticity and stress adaptation~\citep{ALDH}, contributes to metabolic plasticity and oxidative stress resistance~\citep{ALDH3B1}.
% whereas \textit{ALDH3B2} is associated with stemness and metabolic plasticity~\citep{ALDH}.
This axis implies that DCIS/LCIS regions transmit \textit{COL9A2}-driven cues to the tumor margin, inducing stem-like subpopulations with heightened resistance and metastatic potential~\citep{ALDH,SDC1}. Targeting this pathway may provide a strategy to eradicate cancer stem cell–like populations and reduce recurrence risk~\citep{stem_cell}.

In summary, the ligand–target network highlights how early lesions (DCIS/LCIS) drive communication with surrounding regions, providing mechanistic insights into tumor progression and potential intervention points.

\section{Conclusion}
\raggedbottom
In this work, we presented MultiST, a multimodal framework that integrates spatial gene expression and histological images for comprehensive spatial transcriptomics analysis. By combining a graph-regularized gene encoder, a color-normalized image encoder, and a cross-attention fusion module, MultiST effectively captures spatial, morphological, and transcriptional dependencies. The incorporation of GAN-based adversarial regularization and MMD alignment further enhanced the robustness of expression embeddings under sparse and noisy conditions.

Through extensive evaluations on both brain and breast cancer datasets, MultiST demonstrated consistent improvements in clustering accuracy, pseudotime reconstruction, and CCI analysis compared to existing methods. In particular, our results revealed biologically meaningful insights into cortical lamination and tumor microenvironmental communication, underscoring the utility of multimodal integration for resolving tissue heterogeneity.

Looking forward, MultiST provides a generalizable foundation for downstream applications such as cell type deconvolution, and spatial drug-response prediction. We anticipate that this framework will facilitate deeper mechanistic understanding of tissue organization and disease progression, and ultimately support translational efforts in precision oncology and spatial systems biology.

%%%%%%%%%%%%%%

% \begin{appendices}

% \section{Acknowledgements}\label{sec12}%
% We thank the anonymous reviewers for their valuable
% suggestions.

% \section{Supplementary data}\label{sec11}
% Supplementary data are available at Bioinformatics online.
% Conflict of interest: No competing interest is declared.

% \end{appendices}
% Fusce mauris. Vestibulum luctus nibh at lectus. Sed bibendum, nulla a faucibus semper, leo velit ultricies tellus, ac
% venenatis arcu wisi vel nisl. Vestibulum diam. Aliquam pellentesque, augue quis sagittis posuere, turpis lacus congue
% quam, in hendrerit risus eros eget felis. Maecenas eget erat in sapien mattis porttitor. Vestibulum porttitor. Nulla
% facilisi. Sed a turpis eu lacus commodo facilisis. Morbi fringilla, wisi in dignissim interdum, justo lectus sagittis dui, et
% vehicula libero dui cursus dui. Mauris tempor ligula sed lacus. Duis cursus enim ut augue. Cras ac magna. Cras nulla.

% \section{Data availability}\label{sec13}%
% The datasets used in this study are publicly available. 
% the human dorsal lateral prefrontal cortex dataset can be accessed from \url{http://spatial.libd.org/spatialLIBD/}. The human breast cancer dataset can be download from \url{https://github.com/JinmiaoChenLab/SEDR\_analyses/tree/master/data}. The code used for data processing and analysis is available at \url{https://github.com/weiawei1111}. 
% \vspace{-0.48em}
% \vspace{-5pt}
% \bibliographystyle{unsrt}
% \bibliography{reference}

% \vspace{-5pt}  
% \section*{References} 
\begin{small} 
\bibliographystyle{unsrt}
\bibliography{reference}
\end{small}
\end{document}